\documentclass{article}

\usepackage{microtype}
\usepackage{graphicx}
\usepackage{subcaption}
\usepackage{booktabs} %

\usepackage{hyperref}

\usepackage[preprint]{icml2026}

\usepackage{amsmath}
\usepackage{amssymb}
\usepackage{mathtools}
\usepackage{amsthm}

\usepackage[capitalize,noabbrev]{cleveref}

\usepackage{url}
\usepackage{graphicx} 
\usepackage{wrapfig}
\usepackage{booktabs}
\usepackage{multirow}
\usepackage[normalem]{ulem}
\useunder{\uline}{\ul}{}
\usepackage{fontawesome}
\usepackage[table,xcdraw]{xcolor}
\usepackage{pifont}

\usepackage{amsmath,amsfonts,bm}

\def\secref#1{section~\ref{#1}}

\def\eqref#1{equation~\ref{#1}}

\def\1{\bm{1}}

\DeclareMathAlphabet{\mathsfit}{\encodingdefault}{\sfdefault}{m}{sl}
\SetMathAlphabet{\mathsfit}{bold}{\encodingdefault}{\sfdefault}{bx}{n}

\newcommand{\rFID}{\mathrm{rFID}}
\newcommand{\gFID}{\mathrm{gFID}}

\newcommand{\CARD}{\texttt{CARD}}

\newcommand{\modelname}{\texttt{SMAP}}
\newcommand{\Th}[1]{{\scshape #1}}

\theoremstyle{plain}

\theoremstyle{definition}

\theoremstyle{remark}

\usepackage[textsize=tiny]{todonotes}

\icmltitlerunning{Semantic-Aware Prefix Learning for Token-Efficient Image Generation}

\begin{document}

\twocolumn[
  \icmltitle{Semantic-Aware Prefix Learning for Token-Efficient Image Generation}

  \icmlsetsymbol{equal}{*}

  \begin{icmlauthorlist}
    \icmlauthor{Qingfeng Li}{casia,ks}
    \icmlauthor{Haoxian Zhang}{ks}
    \icmlauthor{Xu He}{ks,thu}
    \icmlauthor{Songlin Tang}{ks}
    \icmlauthor{Zhixue Fang}{ks}
    \icmlauthor{Xiaoqiang Liu}{ks}
    \icmlauthor{Pengfei Wan}{ks}
    \icmlauthor{Guoqi Li}{casia}
  \end{icmlauthorlist}

  \icmlaffiliation{casia}{Institute of Automation, Chinese Academy of Sciences}
  \icmlaffiliation{ks}{Kling Team, Kuaishou Technology}
  \icmlaffiliation{thu}{Tsinghua University}

  \icmlcorrespondingauthor{Guoqi Li}{guoqi.li@ia.ac.cn}

  \vskip 0.3in
]

\printAffiliationsAndNotice{%
Qingfeng Li completed this work during an internship at Kling Team, Kuaishou Technology.\ 
Haoxian Zhang is the project leader.
}

\begin{abstract}
Visual tokenizers play a central role in latent image generation by bridging high-dimensional images and tractable generative modeling. 
However, most existing tokenizers are still trained with reconstruction-dominated objectives, which often yield latent representations that are only weakly grounded in high-level semantics. 
Recent approaches improve semantic alignment, but typically treat semantic signals as auxiliary regularization rather than making them functionally necessary for representation learning.
We propose \texttt{SMAP}, a \textbf{S}e\textbf{M}antic-\textbf{A}ware \textbf{P}refix tokenizer that injects class-level semantic conditions into a query-based 1D tokenization framework. 
To make semantics indispensable during training, \texttt{SMAP} introduces a \emph{tail token dropping} strategy, which forces semantic conditions and early latent prefixes to bear increasing responsibility under progressively reduced token budgets. 
To verify that the resulting latent space is useful for generation rather than reconstruction alone, we further introduce \texttt{CARD}, a hybrid \textbf{C}ausal \textbf{A}uto\textbf{R}egressive--\textbf{D}iffusion generator. 
Extensive experiments on ImageNet show that \texttt{SMAP} consistently improves reconstruction quality across discrete and continuous tokenization settings, and that its semantically grounded latent space yields strong downstream generation performance under compact token budgets.
\end{abstract}

\section{Introduction}
\label{sec:intro}

In recent years, image generation has achieved substantial progress across multiple modeling paradigms, including diffusion models~\citep{rombach2022high, yao2024fasterdit, ma2024sit}, autoregressive visual models~\citep{esser2021taming, li2024mar, tian2024var}, and masked generative approaches~\citep{chang2022maskgit, li2023mage}. Despite differences in their generative mechanisms, these methods share a common architectural principle: images are first mapped from the high-dimensional pixel space into a compact latent representation through a learned image encoder or tokenizer~\citep{rombach2022high, esser2021taming, yu2022scaling}. This latent space, which may be continuous or discrete, aims to preserve essential semantic and structural information while significantly reducing dimensionality. 
Existing research has largely focused on improving the generative stage through advances in model architectures~\citep{peebles2023dit} and training objectives~\citep{ma2024sit}, while the role of the latent representation learning mechanism remains comparatively underexplored. However, the structure, expressiveness, and inductive biases of the latent space critically determine both the efficiency and the performance ceiling of downstream generative models~\citep{nextstepteam2025nextstep1, ke2025hyperspherical}, underscoring the importance of systematically studying and improving image encoding and tokenization strategies.

\begin{figure*}[!ht]
    \centering
    \includegraphics[width=0.75\linewidth]{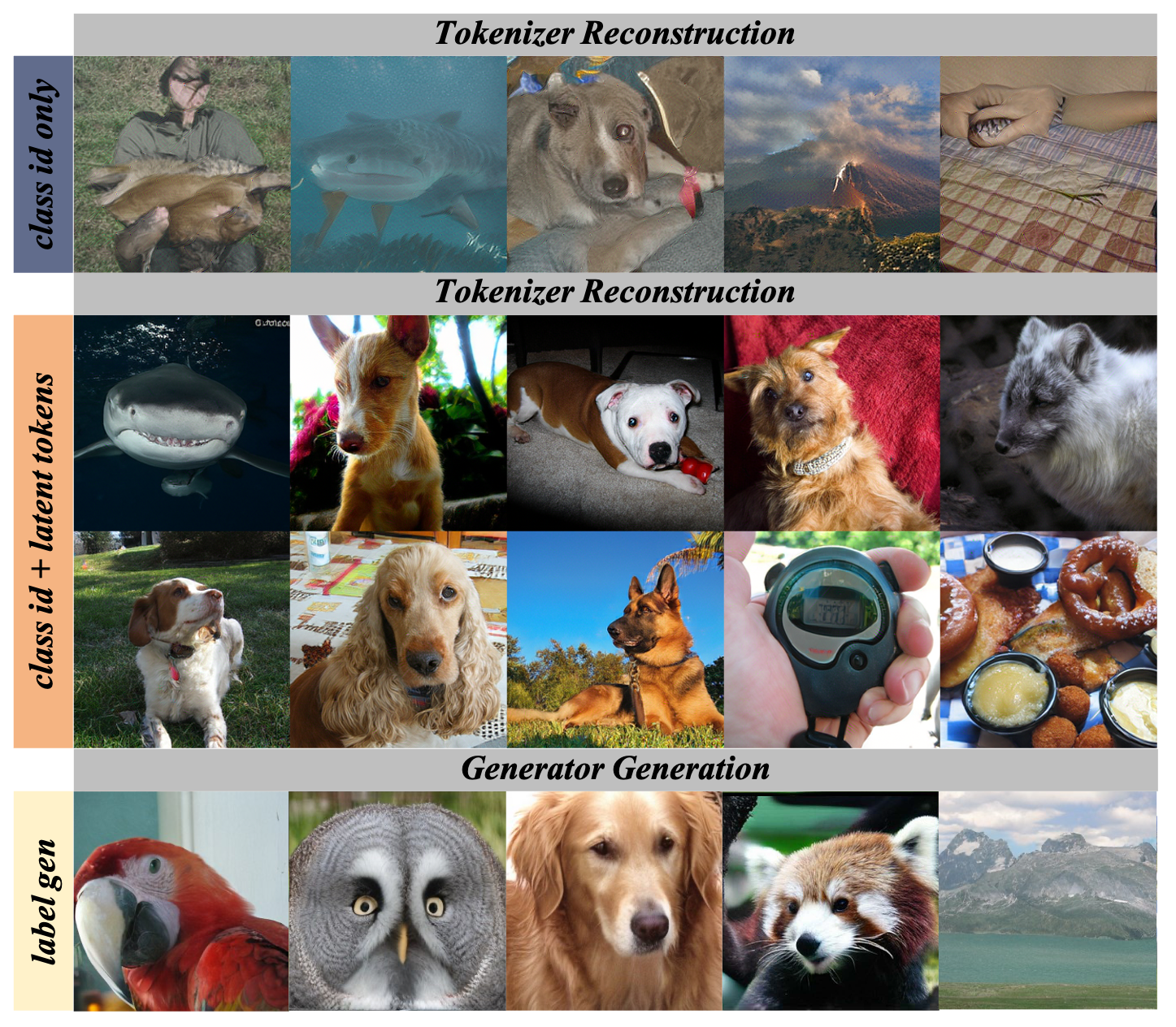}
    \caption{\textbf{Semantic-aware prefix learning in reconstruction and generation.}
    \textbf{Top:} Using only the class condition, \texttt{SMAP} reconstructs images that already capture category-level semantics and coarse global structure.
    \textbf{Middle:} Adding latent tokens substantially improves reconstruction fidelity and restores instance-specific details, showing that semantic conditions and latent prefixes play complementary roles.
    \textbf{Bottom:} Based on the resulting semantically grounded token space, \texttt{CARD} generates high-quality class-conditional images.}
    \vspace{-4mm}
    \label{fig:teaser}
\end{figure*}

Although an increasing body of work has recognized the importance of latent space quality~\citep{kingma2014, tschannen2025givt, rombach2022high}, most existing approaches ~\citep{yu2022vectorquantized, zhu2023designing, yu2024language} still train visual tokenizers using reconstruction-dominated objectives.
Their learned latent or token space may exhibit only weak alignment with high-level concepts, limiting its effectiveness as an interface for downstream generative models and impairing semantic controllability.

To address this mismatch, recent studies have begun to introduce semantic inductive biases into tokenizer pretraining through alignment or regularization strategies~\citep{yu2024titok, kim2025tatitok, yao2025vtp}. A common approach leverages pretrained semantic encoders, such as \texttt{CLIP}~\citep{Radford2021clip}, or representation alignment signals (e.g., \texttt{REPA}~\citep{yu2025repa}), to encourage correlation between latent codes and high-level semantics. 
Along similar lines, Visual Tokenizer Pretraining (\texttt{VTP}~\citep{yao2025vtp}) observes that reconstruction-centric training objectives bias token representations toward low-level visual information and struggle to yield concise semantic abstractions, and consequently advocates injecting semantic signals during tokenizer learning.

More importantly, semantic alignment~\citep{chen2025softvqvae, yu2025repa} alone does not guarantee that the token space can carry and express the high-level structural information required for image generation. 
Many existing approaches merely encourage token representations to be correlated with semantic features in a loose sense, without explicitly requiring semantic signals to bear essential informational responsibility during reconstruction and representation learning. 
We therefore argue that \emph{the core challenge lies not in whether semantics are aligned, but in how semantic information is made an indispensable component of tokenizer pretraining—such that global structure and high-level concepts are encoded into usable and transferable
token representations.}

To this end, we propose \texttt{SMAP}, a semantically aware image tokenizer that encodes high-level semantics as prefix-preserved invariants. By construction, semantic information actively participates in both reconstruction and representation learning throughout pretraining, explicitly driving the tokenizer to encode global structure and high-level concepts. \texttt{SMAP} employs a query-based 1D tokenizer architecture and a principled tail token dropping strategy to learn information-ordered token sequences, enabling length-adaptive representations with strong semantic grounding. We refer to this behavior as semantic-aware prefix learning: semantic conditions encode high-level identity in the prefix, while later latent tokens progressively refine instance-level detail.

Building upon \texttt{SMAP}, we further propose \texttt{CARD}, a class of hybrid autoregressive–diffusion generative models for image generation. Following the staged generation principle of \texttt{MAR}~\citep{li2024mar}, \texttt{CARD} decomposes image generation into two complementary components: an autoregressive module that models high-level structural dependencies in the latent space, followed by a \texttt{Flow Matching}~\citep{lipman2023flow}–based continuous density model that captures and refines the conditional distribution, enabling high-quality image synthesis.

In brief, our contributions are threefold.
\vspace{-3mm}
\begin{itemize}
    \item We identify a central limitation of existing tokenizer training pipelines: semantics are typically encouraged through loose alignment objectives, rather than being made functionally necessary for reconstruction and representation learning.
    \vspace{-2mm}
    \item We propose \texttt{SMAP}, a semantic-aware 1D tokenizer that incorporates semantic conditions as prefix-preserved invariants and enforces semantic dependency through token truncation, resulting in semantically grounded, information-ordered, and length-adaptive token sequences.
    \vspace{-2mm}
    \item We develop \texttt{CARD}, a hybrid autoregressive--diffusion generator built on top of \texttt{SMAP}, and demonstrate that semantically grounded tokenization consistently improves both tokenizer reconstruction and downstream image generation under compact token budgets.
\end{itemize}

\section{Related Work}
\label{sec:related}

\noindent \textbf{Image Tokenization.}
Modern generative image models rely critically on image tokenization to enable efficient and scalable generation~\citep{esser2021taming,rombach2022high,chang2022maskgit,yu2022scaling}.
By encoding images into discrete~\citep{oord2017vq, ryu2024vqgan} or continuous~\citep{rombach2022high}
latent tokens, these models avoid operating directly in pixel space
and instead focus on learning semantically meaningful representations.
Early work used autoencoders~\citep{hinton2006reducing,vincent2008extracting} to learn low-dimensional latent representations, which were later extended to structured generative models such as VAEs and VQ-GAN~\citep{van2017neural,razavi2019generating,esser2021taming}.
VQ-GAN–style~\citep{goodfellow2014generative, esser2021taming, yu2021vector, zheng2023online, yu2024language} discrete formulations naturally align with autoregressive~\citep{esser2021taming}
and masked generative models~\citep{chang2022maskgit}, facilitating the adoption of
techniques originally developed for language modeling~\citep{gpt3}.
Continuous tokenization follows the variational autoencoder (VAE) framework~\citep{kingma2013vae},
in which latent representations are modeled as samples from a normal distribution.

\noindent \textbf{Image Generation.}
Image generation methods are predominantly categorized into autoregressive and diffusion models.
Early autoregressive approaches were primarily built upon convolutional neural networks~\citep{van2016conditional}, and were later extended with Transformer-based architectures~\citep{vaswani2017attention,lee2022rqvae,liu2024customize,sun2024llama,yu2025randomized} to improve scalability and modeling capacity~\citep{chang2022maskgit,tian2024var}.
Diffusion models have demonstrated strong generative performance since their introduction~\citep{sohldickstein2015deepunsupervisedlearningusing}. Subsequent developments refined the denoising process and significantly improved sample quality~\citep{nichol2021ddpm,dhariwal2021diffusion,song2022ddim}.
A pivotal advance in both performance and efficiency was achieved by latent diffusion models~\citep{vahdat2021scorelatentdiffusion,rombach2022ldm},
which leverage learned tokenizers to perform denoising in a compact latent space,
thereby reducing computational cost while preserving visual fidelity ~\citep{van2017neural,esser2021taming,peebles2023dit,qiu2025robust}.
Recent research has further advanced image generation by improving tokenizer design~\citep{chen2025softvqvae,zha2024language,yao2025reconstruction} and by exploring hybrid frameworks that combine diffusion and autoregressive modeling paradigms~\citep{li2024mar}.

\section{Method}
\label{sec:method}

This section presents our method. We first review query-based 1D tokenization for latent image modeling in both discrete and continuous settings (\secref{sec:prelim}). 
We then introduce \texttt{SMAP}, a semantic-aware tokenizer that incorporates conditional semantics directly into token formation and reconstruction (\secref{sec:SMAP}). 
Finally, we present \texttt{CARD}, a hybrid autoregressive--diffusion generator designed to exploit the information ordering induced by \texttt{SMAP} (\secref{sec:CARD}).

\subsection{Preliminary: Query-Based 1D Tokenization}
\label{sec:prelim}

Recent token-based image representations increasingly adopt a query-based tokenization paradigm, drawing inspiration from \texttt{Q-Former}-style architectures ~\citep{li2023blip2, yu2024titok, li2024imagefolder, chen2025softvqvae}, in which a fixed set of learnable queries selectively attends to visual features to extract compact representations.
Several recent visual tokenizers adopt this query-based formulation, among which \texttt{TiTok} is a representative example.

\texttt{TiTok} is a transformer-based, one-dimensional vector-quantized (\texttt{VQ}) tokenizer that departs from conventional grid-structured latent representations. Instead of preserving a two-dimensional spatial layout, \texttt{TiTok} represents an image using a compact sequence of latent tokens.
Given an input image $\mathbf{I} \in \mathbb{R}^{H \times W \times 3}$, \texttt{TiTok} first applies a patch embedding operation with downsampling factor $f$, producing visual patch features $\mathbf{F} \in \mathbb{R}^{(\frac{H}{f} \times \frac{W}{f}) \times D}$. 
A set of learnable latent tokens $\mathbf{L} \in \mathbb{R}^{K \times D}$ is then concatenated with the patch tokens along the sequence dimension. The resulting sequence is processed by a Vision Transformer (\texttt{ViT}) encoder $\mathtt{Enc}$ to produce token embeddings, from which only the embeddings corresponding to the latent tokens are retained:
\begin{align}
    \label{eq:enc}
    [\_ ; \mathbf{Z}_{\mathrm{1D}}] = \mathtt{Enc}([\mathbf{F} ; \mathbf{L}]).
\end{align}
where $[\cdot;\cdot]$ denotes concatenation along the sequence dimension, $\mathbf{Z}_{\mathrm{1D}} \in \mathbb{R}^{K \times D}$ represents the resulting one-dimensional latent tokens, and $\_$ denotes tokens that are discarded in subsequent processing.

The resulting one-dimensional latent tokens $\mathbf{Z}_{\mathrm{1D}} \in \mathbb{R}^{K \times D}$ can be instantiated using discrete or continuous representations.
In the original \texttt{TiTok} framework, latent tokens are quantized using a vector quantizer $\mathtt{Quant}(\cdot)$, which maps each token to its nearest entry in a learnable codebook.
Subsequent work ~\citep{kim2025tatitok} extends this formulation by modeling latent tokens as continuous random variables and applying variational regularization, producing a compact 1D VAE representation that avoids information loss induced by quantization.
For notational convenience, we use a unified regularization operator $\mathtt{Regu}(\cdot)$ to denote the latent regularization applied before decoding. Its concrete instantiation under the VQ, KL, and SoftVQ formulations is provided in~\autoref{app:regu}.

During de-tokenization, a sequence of mask tokens $\mathbf{M} \in \mathbb{R}^{(\frac{H}{f} \times \frac{W}{f}) \times D}$ is introduced and concatenated with the latent tokens, regardless of whether they are discrete or continuous.
The combined sequence is then passed through a Vision Transformer decoder
$\mathtt{Dec}$ to reconstruct the image \(\mathbf{\hat{I}}\):
\begin{align}
    \label{eq:dec}
    [\_~; \hat{\mathbf{I}}] =
    \mathtt{Dec}([\mathtt{Regu}(\mathbf{Z}_{\mathrm{1D}}) ; \mathbf{M}]).
\end{align}

\subsection{SMAP: Semantic-Aware Prefix Tokenization with Semantics-Preserved Prefixes}
\label{sec:SMAP}

\noindent \textbf{Overall Design.}
\texttt{SMAP} is designed to force semantic information to become a functional prefix-level carrier of reconstruction, rather than an auxiliary alignment target.
To this end, \texttt{SMAP} extends query-based 1D tokenization with two key ideas.
First, semantic conditions are injected into both the encoder and decoder as explicit sequence elements, allowing semantic cues to participate in token formation and reconstruction. 
Second, a tail token dropping strategy is applied during training so that semantic conditions and early token prefixes must progressively absorb more global structural responsibility.
Together, these two mechanisms encourage the tokenizer to learn information-ordered latent sequences with strong semantic grounding.

\noindent \textbf{Query-based Encoder–Decoder Formulation.}
Given an input image $\mathbf{I}$, \texttt{SMAP} first extracts visual features
$\mathbf{F}$ using a \texttt{ViT}-based image encoder.
Similar to \texttt{TiTok}, a set of learnable latent queries
$\{\mathbf{q}_i\}_{i=1}^{K}$ is used to aggregate visual information via
self-attention, producing the output token sequence
$\mathbf{z}^{[t]}_{1:K}$ from the $t$-th block.
Similarly, in the decoder, information is propagated through self-attention between the learnable mask tokens $\{\mathbf{m}_i\}_{i=1}^{L}$ and the latent tokens $\{\mathbf{\hat{q}}_i\}_{i=1}^{K}$.
Unlike \texttt{TiTok}, \texttt{SMAP} directly reconstructs the final image \(\mathbf{\hat{I}}\) from the mask tokens $\{\mathbf{m}_i\}_{i=1}^{K}$ 
in the output of the decoder, rather than using latent tokens as primary reconstruction carriers.
This design assigns latent tokens the role of encoding semantic and global information, while explicitly delegating spatially structured image synthesis to the mask tokens.

\begin{figure*}[!ht]
    \centering
    \includegraphics[width=0.85\linewidth]{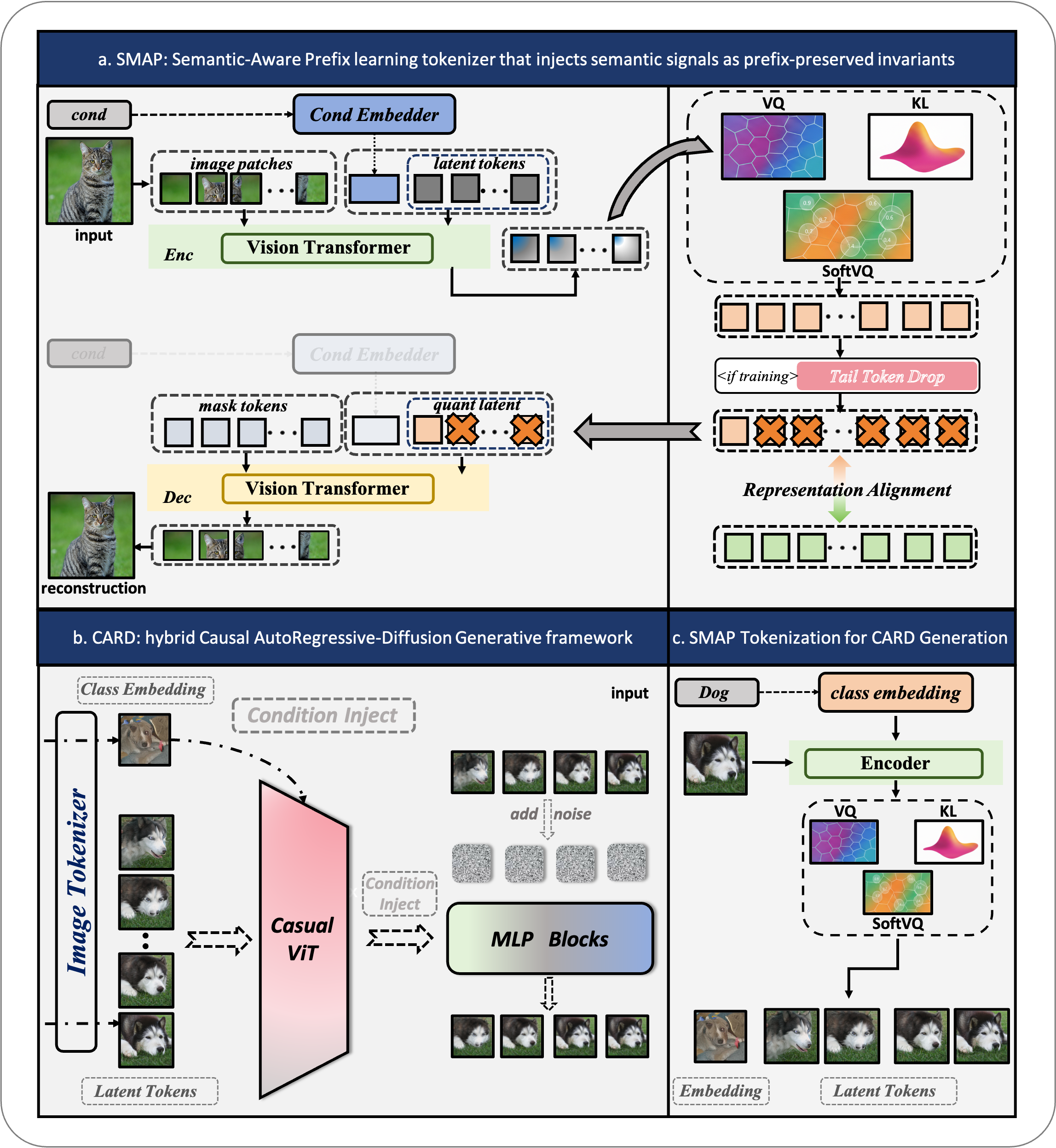}
    \caption{
    Overview of our method.
    \textbf{(a)} proposes a novel mechanism for semantic injection. It extracts conditional embeddings from class labels and inserts them between visual patch tokens and learnable latent tokens. The condition embeddings act as intermediaries that interact jointly with image patches to guide the formation of latent tokens. It further strengthens semantic dependency through a tail token dropping strategy.
    \textbf{(b)} proposes a hybrid Causal AutoRegressive–Diffusion framework that fully leverages \texttt{SMAP}’s capabilities.
    \textbf{(c)} shows the \texttt{SMAP} tokenization process for \texttt{CARD} generation.
    }
    \vspace{-4mm}
    \label{fig:pipeline}
\end{figure*}

\begin{figure*} [!]

{\makebox[\textwidth][c]{\includegraphics[width=\textwidth,trim={ 0cm 0cm 0cm 0cm},clip]{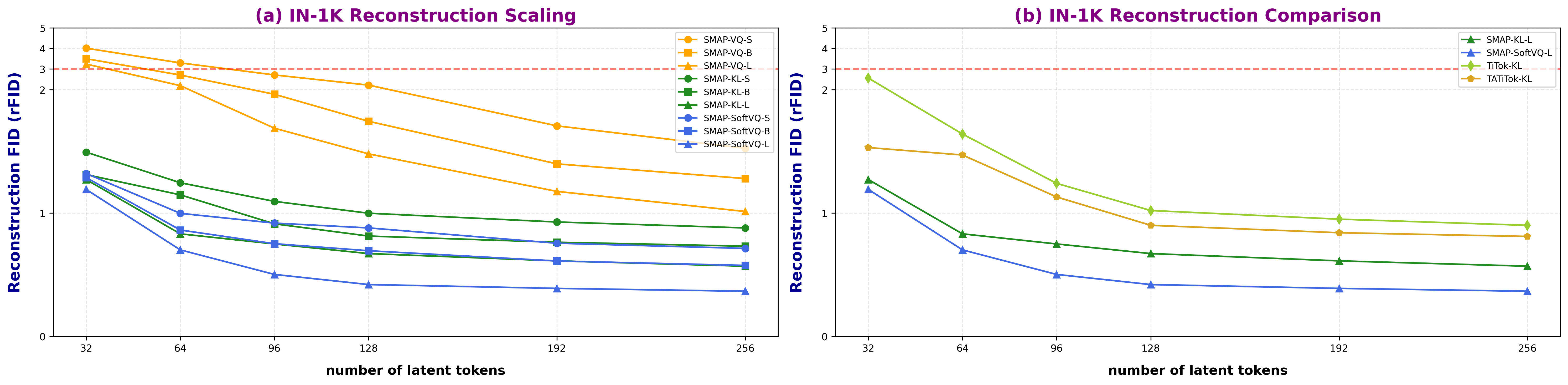}}}
    \caption{\textbf{ImageNet-1K reconstruction scaling and comparison.}
    \textbf{(a)} Reconstruction FID ($\rFID$) of \texttt{SMAP} under different token budgets and model scales.
    Across VQ, KL, and SoftVQ variants, increasing the number of latent tokens consistently improves reconstruction quality, and larger \texttt{SMAP} models achieve stronger performance under the same token budget.
    \textbf{(b)} Reconstruction comparison with prior 1D tokenizers.
    At matched token lengths, \texttt{SMAP} consistently outperforms \texttt{TiTok} and \texttt{TA-TiTok}, with the largest gains observed in continuous latent settings.
    Overall, the results show that \texttt{SMAP} scales favorably with both token budget and model capacity, while providing substantially better reconstruction quality than existing baselines.}
    \vspace{-4mm}
    \label{fig:ablation_experimental}
\end{figure*}

Moreover, \texttt{SMAP} supports both discrete and continuous forms of token regularization within a unified tokenizer framework, enabling flexible training across different generative paradigms such as autoregressive and diffusion models.
By jointly designing the query-based tokenization mechanism and the ViT-based encoder–decoder architecture, \texttt{SMAP} can more fully exploit the scaling properties of \texttt{Transformer} models.
\autoref{fig:ablation_experimental}(a) illustrates the scaling behavior of \texttt{SMAP}, while~\autoref{fig:ablation_experimental}(b) compares the current design against prior tokenizers.
In contrast to previous approaches~\citep{yu2024titok, miwa2025onedpiece} that rely on multi-stage optimization or additional pretraining procedures, our tokenizer significantly simplifies the training pipeline in a single stage while maintaining strong reconstruction quality and scalability to large datasets and model sizes.

\noindent \textbf{Semantic Injection Mechanism.}
Although methods such as \texttt{REPA} encourage tokenizer representations to correlate with semantic features to some extent, they typically treat semantic signals as auxiliary alignment or regularization objectives, without explicitly requiring semantics to bear essential informational responsibility during reconstruction and representation learning. 
To make semantic information an indispensable component of tokenizer pre-training, we introduce an explicit semantic injection mechanism.
We first discuss the construction of conditional embeddings
$\mathbf{C} \in \mathbb{R}^{N \times D}$ from semantic supervision.
For class-level conditions, we jointly train an additional class embedding module within the tokenizer, whose embedding dimensionality is aligned with that of the learnable latent tokens, allowing direct concatenation and interaction along the sequence dimension.

As shown in~\autoref{fig:pipeline}(a), we derive conditional embeddings 
$\mathbf{C} \in \mathbb{R}^{(N \times D)}$
from class labels and insert them between visual patch tokens $\mathbf{V} \in \mathbb{R}^{(L \times D)}$
and learnable latent queries $\mathbf{L} \in \mathbb{R}^{(K \times D)}$. 
The resulting token sequence is then jointly processed by the encoder
$\mathtt{Enc}$, allowing visual content and explicit semantic cues to interact through self-attention and jointly shape the formation of latent token representations.
\begin{align}
    \label{eq:SMAP-enc}
    [\_~; \_~; \mathbf{Z}_{\mathrm{1D}}] = \mathtt{Enc}([\mathbf{V} ; \mathbf{C} ; \mathbf{L}]).
\end{align}
In the de-tokenization stage, \texttt{SMAP} symmetrically incorporates the semantic embeddings introduced during tokenization.
As illustrated in the de-tokenization module in~\autoref{fig:pipeline}(a), the conditional embeddings are injected between the learnable mask tokens 
$\mathbf{M} \in \mathbb{R}^{L \times D}$ 
and the processed latent tokens
$\mathbf{\hat{L}} \in \mathbb{R}^{K \times D}$, 
and jointly modeled through the decoder $\mathtt{Dec}$, enabling the mask tokens to aggregate the information required for reconstruction and ultimately generate the image $\mathbf{\hat{I}}$.
\begin{align}
    \label{eq:SMAP-dec}
    [\mathbf{\hat{I}} ; \_~; \_~;] = \mathtt{Dec}([\mathbf{M} ; \mathbf{C} ; \mathbf{\hat{L}}]).
\end{align}

\noindent \textbf{Tail token dropping.}
To enforce semantic dependency during tokenizer pre-training, we introduce a
\emph{tail token dropping} strategy that perturbs the latent token sequence at training time.
Let the encoder output latent tokens be $\mathbf{Z}^{\mathtt{1D}}_{1 : K}$.
At each iteration, we sample a retained prefix length 
$k \in \{0,1,\dots,K\}$ and keep only the prefix tokens $\mathbf{Z}^{\mathtt{1D}}_{1:k}$,
while removing the tail tokens $\mathbf{Z}^{\mathtt{1D}}_{k+1:K}$ (or equivalently masking them out in the attention computation).
The extreme case $k=0$ corresponds to dropping all latent tokens, in which case the decoder must reconstruct the image $\mathbf{\hat{I}}$ using only the conditional embeddings
$\mathbf{C} \in \mathbb{R}^{N \times D}$ together with the mask tokens $\mathbf{M} \in \mathbb{R}^{K \times D}$.
This training-time perturbation explicitly increases the informational burden placed on the conditional embeddings.
Importantly, this strategy operates \emph{directly on the token sequence}, so we can construct the decoder input by concatenating the retained latent prefix $\mathtt{Regu}(\mathbf{Z}^{\mathtt{1D}}_{1:k})$
with the semantic embeddings $\mathbf{C}$
and the learnable mask tokens $\mathbf{M}$.
\begin{align}
    [\mathbf{\hat{I}} ; \_~; \_~;] = 
    \mathtt{Dec}([\mathbf{M} ; \mathbf{C} ; \mathtt{Regu}(\mathbf{Z}_{1:k})])
\end{align}
The prefix length $k$ is sampled only during training.
Specifically, we draw $k$ from a uniform distribution over token indices $k \sim \mathtt{Unif}\{0, 1,\dots, K\}$, 
so that different token budgets are randomly explored across training iterations.
Consequently, the semantic prefix becomes the only information pathway that is preserved across all sampled token budgets.

\subsection{CARD: Hybrid Diffusion--Autoregressive Generative Model}
\label{sec:CARD}

\noindent \textbf{Architecture.}
To fully exploit the semantic-aware and information-ordered latent space learned by \texttt{SMAP}, we propose \texttt{CARD}, a hybrid generative framework that combines causal autoregressive modeling with diffusion-style refinement.
As illustrated in~\autoref{fig:pipeline}(b), \texttt{CARD} first applies a causal transformer to model the structural dependencies among latent tokens in an autoregressive manner, thereby capturing coarse global structure and long-range token interactions.
The autoregressive predictions are then passed to a lightweight continuous refinement module, instantiated as a stack of \texttt{MLP} blocks, which denoises noisy latent variables and improves generation fidelity.

Concretely, let $\mathbf{Z}_{\mathrm{1D}}$ denote the latent token sequence.
The causal autoregressive module produces structure-aware latent predictions
$\mathtt{AR}(\mathbf{Z}_{\mathrm{1D}})$, which are used as conditional inputs to the refinement model.
Given a noisy latent $\mathbf{x}_t$ at timestep $t$, the denoising velocity is predicted as
\begin{equation}
\mathbf{v}_t = \mathtt{MLP}\bigl(\mathbf{x}_t,\, t,\, \mathtt{AR}(\mathbf{Z}_{\mathrm{1D}})\bigr),
\end{equation}
where $\mathbf{x}_t$ is the noisy latent variable and $\mathtt{AR}(\cdot)$ denotes the autoregressive outputs.
In contrast to \texttt{MAR}~\citep{li2024mar}, which directly concatenates condition embeddings with image tokens, \texttt{CARD} injects conditions into the generator through adaptive normalization, following the conditioning strategy of \texttt{DiT}~\citep{peebles2023dit}.
This design preserves the compact token structure while enabling flexible conditional control.

\noindent \textbf{Semantic Condition Sharing.}
A key design choice of \texttt{CARD} is that its conditioning signal is not introduced through a separately trained class encoder.
Instead, as illustrated in~\autoref{fig:pipeline}(c), we directly reuse the class-aware semantic embedding learned during \texttt{SMAP} pretraining as the condition input for generation.
More specifically, the class label is first mapped to a semantic embedding by the same condition embedding module used in the tokenizer, and this embedding is then paired with the latent tokens produced by \texttt{SMAP}.
The resulting shared semantic space is subsequently used throughout \texttt{CARD}, ensuring that the condition signal used in generation is consistent with the semantic prefix that shaped tokenizer learning.

This semantic condition sharing has two advantages.
First, it removes the need to introduce an additional condition encoder on the generator side, thereby simplifying the overall architecture.
Second, it strengthens semantic consistency between tokenization and generation: the same embedding space that guides semantic prefix learning in \texttt{SMAP} is also used to control downstream image synthesis in \texttt{CARD}.
Empirically, this sharing mechanism improves the alignment between class conditions and generated content, and further demonstrates that the semantic representations learned by \texttt{SMAP} are transferable and functionally useful for downstream generation. Detailed empirical analysis of this design is provided in Section~\autoref{exp}.

\begin{figure*}[!ht]
    \centering
    \includegraphics[width=0.85\linewidth]{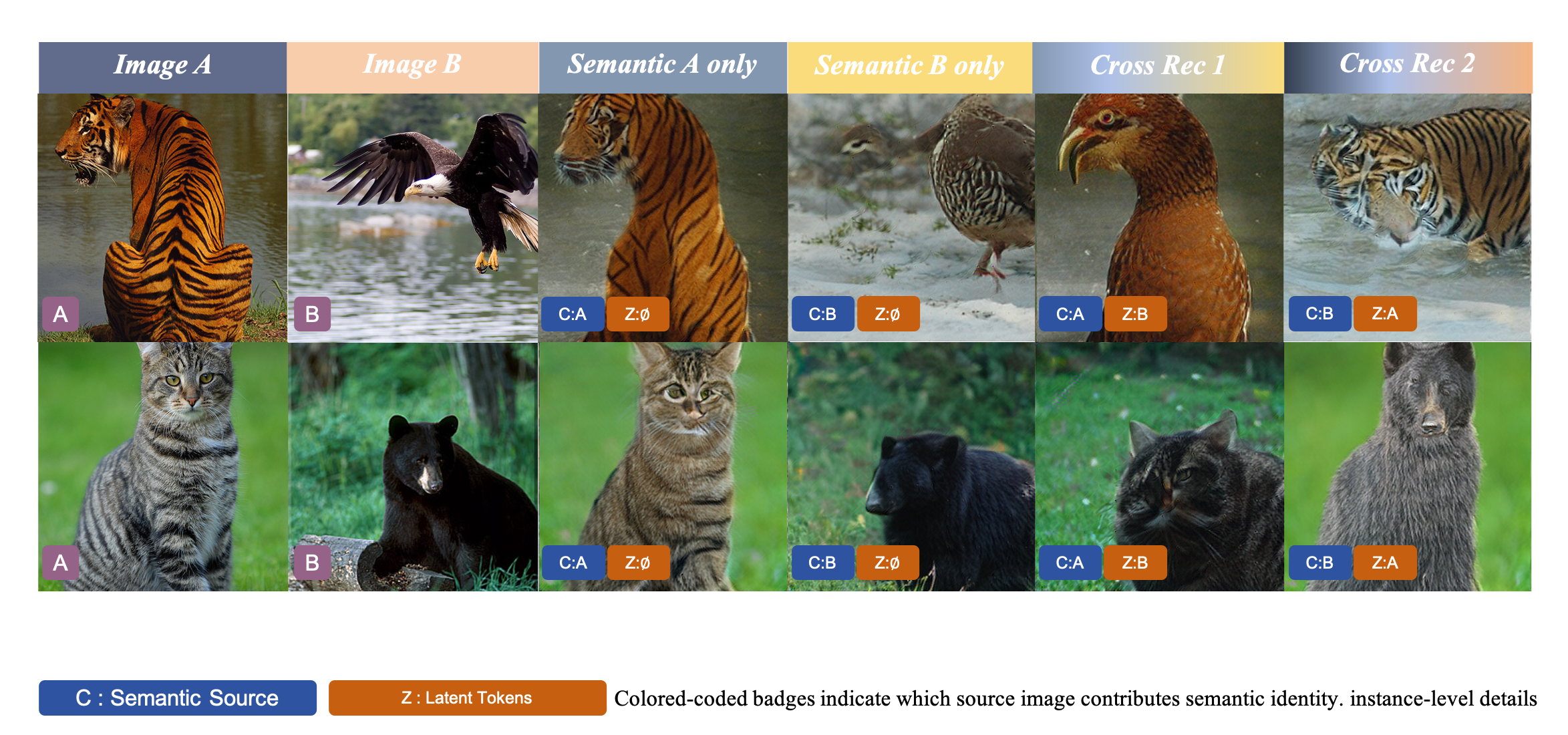}
    \caption{
    \textbf{Semantic identity is controlled by $C$, while instance-level details are carried by $Z$.}
    We visualize reconstructions obtained by independently manipulating the semantic condition $C$ and latent tokens $Z$.
    Using only $C$ with $Z=\emptyset$ yields coarse reconstructions that preserve category-level semantics.
    In contrast, cross-combining $C$ from one image with $Z$ from another transfers semantic identity and instance-specific appearance in a complementary manner.}
    \vspace{-4mm}
    \label{fig:ablation_cross}
\end{figure*}

\section{Experiments}
\label{exp}

\subsection{Experiments Setup}
\label{sec:details}

\noindent \textbf{Implementation Details of Tokenizer.}
We use \texttt{SoftVQ} codebase ~\citep{chen2025softvqvae} to train \texttt{SMAP}.
We instantiate three variants of \texttt{SMAP}, all sharing the same encoder–decoder architecture but differing in model scale, with parameter counts of 185M, 391M, and 568M, corresponding to the \texttt{SMAP-S}, \texttt{SMAP-B}, and SMAP-L configurations, respectively.
We consider three latent regularization schemes: \texttt{VQ}~\citep{oord2017vq, yu2022vectorquantized}, \texttt{SoftVQ}~\citep{chen2025softvqvae}, and \texttt{KL}~\citep{takahashi2019variational}.
For the \texttt{VQ} variant, we adopt a codebook size of 8192 with a channel dimension of 64, and train models with latent token lengths of 64 and 128 to align with the settings used in \texttt{TiTok}.
For the \texttt{KL}-based variant, following the design of \texttt{MAR}~\citep{li2024mar}, we model latent representations as continuous features with 16 channels, and consider latent token lengths of 128 and 256.
For the \texttt{SoftVQ} variant, we employ a hierarchical codebook
design~\citep{li2024imagefolder} with four levels and a total codebook size of 8192, while keeping the channel dimension consistent with the KL variant (\emph{i.e.}, 16 channels).
For ablation studies, we additionally evaluate smaller token budgets (32, 64, 96 and 128) to analyze the effect of compact latent representations. For the main comparison and generator experiments, we use 128 tokens unless otherwise specified.
Please refer to~\autoref{app:implementation_details} for additional experimental details.

\noindent \textbf{Implementation Details of Generator.}
For the discrete variants of \texttt{SMAP}, we adopt \texttt{LlamaGen}~\citep{sun2024llama} as the generative model to evaluate generation performance, following standard practice for discrete token-based representations.
For the continuous variants of \texttt{SMAP}, we employ our proposed
\texttt{CARD} as the generator, which is specifically designed to match the inductive bias and information ordering induced by \texttt{SMAP}.
We consider three variants of \texttt{CARD}, namely \texttt{CARD-B},
\texttt{CARD-L}, and \texttt{CARD-XL}, with 234M, 568M, and 1.1B parameters, respectively.
Detailed architectural configurations for each variant are provided in
\autoref{tab:CARD_configuration}.

\begin{table}[!ht]
    \caption{\textbf{Architecture Configuration of CARD.} Following \texttt{MAR}, we scale up blocks across three configurations.}
    \label{tab:CARD_configuration}
    \small
    \centering
    \setlength{\tabcolsep}{0.45em}
    \begin{tabular}{lcccccc}
      \toprule
      Model  & Depth & Width & Heads & $D_{\text{mlp}}$ & $W_{\text{mlp}}$ & \#Params  \\
      \midrule
      CARD-B & 24 & 768 & 12 & 6 & 1024 & 234M \\
      CARD-L & 32 & 1024 & 16 & 8 & 1280 & 568M \\
      CARD-XL & 40 & 1280 & 16 & 12 & 1536 & 1.1B \\
      \bottomrule
    \end{tabular}
    \vspace{-3mm}
\end{table}

\noindent \textbf{Evaluation Metrics.}
Our evaluation protocol closely follows prior work~\citep{yu2022scaling}.
For the reconstruction evaluation of \texttt{SMAP}, we report reconstruction Frechet Inception Distance (FID)~\citep{fid} and Inception Score (IS)~\citep{is} on the ImageNet~\citep{deng2009imagenet} validation set, providing a comprehensive evaluation of reconstruction fidelity and perceptual quality.
To avoid ambiguity, we explicitly distinguish generation and reconstruction
FID throughout the paper, denoted as $\gFID$ and $\rFID$, respectively.
To evaluate generative performance, we train \texttt{CARD} on the latent representations produced by each variant of \texttt{SMAP}. We report $\gFID$ and IS computed over $50{,}000$ generated samples, following the evaluation protocol of \texttt{ADM}~\citep{dhariwal2021adm}.
Detailed experimental results are provided in~\autoref{app:detailed_preliminary}.

\begin{table}[t]
  \caption{\textbf{Ablation on the improved one-stage training recipe.}
  All models are trained and evaluated on \texttt{ImageNet256}. We compare the baseline tokenizer (\texttt{TiTok}) against \texttt{SMAP} under matched token budgets. Numbers in parentheses indicate the change relative to the baseline. Lower $\rFID$ and higher IS are better.}
  \label{tab:ablation-stage-table}
  \centering
  \small
  \setlength{\tabcolsep}{4pt}
  \begin{tabular}{ccc|cc|cc}
      \toprule
      \multirow{2}{*}{arch} & \multicolumn{2}{c|}{tokens} & \multicolumn{2}{c|}{\textbf{TiTok}} & \multicolumn{2}{c}{\textbf{SMAP}} \\
       & \# & c & $\rFID\downarrow$ & IS$\uparrow$ & $\rFID\downarrow$ & IS$\uparrow$ \\
      \midrule
      \multirow{3}{*}{VQ}
      & 32  & -- & 7.72 & 98.3  & 3.24 {\color{blue}(-4.48)} & 230.4 {\color{blue}(+132.1)} \\
      & 64  & -- & 4.25 & 138.0 & 2.29 {\color{blue}(-1.96)} & 260.1 {\color{blue}(+122.1)} \\
      & 128 & -- & 2.63 & 168.1 & 1.47 {\color{blue}(-1.16)} & 280.5 {\color{blue}(+112.4)} \\
      \midrule
      \multirow{3}{*}{KL}
      & 32  & 16 & 2.56 & 171.7 & 1.07 {\color{blue}(-1.49)} & 280.4 {\color{blue}(+108.7)} \\
      & 64  & 16 & 1.64 & 198.0 & 0.96 {\color{blue}(-0.68)} & 299.5 {\color{blue}(+101.5)} \\
      & 128 & 16 & 1.02 & 209.7 & 0.75 {\color{blue}(-0.27)} & 308.9 {\color{blue}(+99.2)} \\
      \bottomrule
  \end{tabular}
  \vspace{-4mm}
\end{table}

\begin{table}[t]
  \caption{\textbf{Ablation on progressive token truncation.}
  All models are trained and evaluated on \texttt{ImageNet256}. We compare \texttt{SMAP} trained without and with progressive token truncation. Numbers in parentheses indicate the change relative to the corresponding model trained without truncation. Lower $\rFID$ and higher IS are better.}
  \label{tab:ablation-generation-table}
  \centering
  \small
  \setlength{\tabcolsep}{4pt}
  \begin{tabular}{ccc|cc|cc}
      \toprule
      \multirow{2}{*}{arch} & \multicolumn{2}{c|}{tokens} & \multicolumn{2}{c|}{\texttt{SMAP} w/o trunc.} & \multicolumn{2}{c}{\texttt{SMAP} w/ trunc.} \\
       & \# & c & rFID$\downarrow$ & IS$\uparrow$ & rFID$\downarrow$ & IS$\uparrow$ \\
      \midrule
      \multirow{3}{*}{VQ}
      & 32  & -- & 3.41 & 220.8 & 3.24 {\color{blue}(-0.17)} & 230.4 {\color{blue}(+9.6)} \\
      & 64  & -- & 2.50 & 253.4 & 2.29 {\color{blue}(-0.21)} & 260.1 {\color{blue}(+6.7)} \\
      & 128 & -- & 1.65 & 269.8 & 1.47 {\color{blue}(-0.18)} & 280.5 {\color{blue}(+10.7)} \\
      \midrule
      \multirow{3}{*}{KL}
      & 32  & 16 & 1.20 & 269.1 & 1.07 {\color{blue}(-0.13)} & 280.4 {\color{blue}(+11.3)} \\
      & 64  & 16 & 1.03 & 278.6 & 0.96 {\color{blue}(-0.07)} & 299.5 {\color{blue}(+20.9)} \\
      & 128 & 16 & 0.75 & 298.4 & 0.69 {\color{blue}(-0.06)} & 308.9 {\color{blue}(+10.5)} \\
      \bottomrule
  \end{tabular}
  \vspace{-4mm}
\end{table}

\subsection{Optimized Image Tokenization with SMAP}

\noindent\textbf{Improved One-Stage Training Recipe.}
\autoref{tab:ablation-stage-table} summarizes the performance gains of our improved one-stage training recipe over the original schemes in~\citep{yu2024titok}.
We observe that the proposed one-stage training consistently outperforms the original \texttt{TiTok} in both the \texttt{VQ} and \texttt{KL} variants, yielding a uniformly lower \Th{rFID} in all evaluated token lengths.
This shows that the improvement is not tied to a specific latent formulation or token budget, but reflects a more robust and effective tokenizer training strategy.

\begin{table*}[t]
\caption{
\textbf{System-level comparison on ImageNet 256$\times$256 conditional generation.}
\texttt{SMAP}+\texttt{CARD} achieves competitive performance under a compact 128-token budget across both KL and SoftVQ variants.
``Model (G)'' denotes the generator, ``\# Params (G)'' its parameter count, ``Model (T)'' the tokenizer, ``\# Params (T)'' its parameter count, and ``\# Tokens'' the number of latent tokens used during generation.
$^\dagger$ indicates that the model was trained on data beyond ImageNet.}
\label{tab:main_256}
\centering
\setlength{\tabcolsep}{4pt}
\renewcommand{\arraystretch}{1.05}

\resizebox{0.9\textwidth}{!}{
\begin{tabular}{l c c c c c c c c c}
\toprule
\multirow{2}{*}{Model (G)} &
\multirow{2}{*}{\# Params (G)} &
\multirow{2}{*}{Model (T)} &
\multirow{2}{*}{\# Params (T)} &
\multirow{2}{*}{\# Tokens$\downarrow$} &
\multirow{2}{*}{rFID$\downarrow$} &
\multicolumn{2}{c}{w/o CFG} &
\multicolumn{2}{c}{w/ CFG} \\
\cmidrule(lr){7-8} \cmidrule(lr){9-10}
& & & & & & gFID$\downarrow$ & IS$\uparrow$ & gFID$\downarrow$ & IS$\uparrow$ \\
\midrule

\multicolumn{10}{l}{\textit{Auto-regressive}} \\
VQGAN~\citep{esser2021taming} & 1.4B & VQ & 23M & 256 & 7.94 & -- & -- & 5.20 & 290.3 \\
ViT-VQGAN~\citep{yu2021vector} & 1.7B & VQ & 64M & 1024 & 1.28 & 4.17 & 175.1 & -- & -- \\
LlamaGen-3B~\citep{sun2024llama} & 3.1B & VQ & 72M & 576 & 2.19 & -- & -- & 2.18 & 263.3 \\
TiTok-S-128~\citep{yu2024titok} & 287M & VQ & 72M & 128 & 1.61 & -- & -- & 1.97 & 281.8 \\
VAR~\citep{tian2024var} & 2B & MSRQ$^\dagger$ & 109M & 680 & 0.90 & -- & -- & 1.92 & 323.1 \\
MAR-H~\citep{li2024mar} & 943M & KL & 66M & 256 & 1.22 & 2.35 & 227.8 & 1.55 & 303.7 \\
\midrule

\multicolumn{10}{l}{\textit{Diffusion-based}} \\
LDM-4~\citep{vahdat2021scorelatentdiffusion} & 400M & KL$^\dagger$ & 55M & 4096 & 0.27 & 10.56 & 103.5 & 3.60 & 247.7 \\
MDTv2-XL/2~\citep{sahoo2024mdtv2} & 676M & -- & -- & -- & -- & 5.06 & 155.6 & 1.58 & 314.7 \\
DiT-XL/2~\citep{peebles2023dit} & 675M & -- & -- & -- & -- & 9.62 & 121.5 & 2.27 & 278.2 \\
SiT-XL/2~\citep{ma2024sit} & 675M & -- & -- & -- & -- & 8.30 & 131.7 & 2.06 & 270.3 \\
\quad + REPA~\citep{yao2024fasterdit} & 675M & -- & -- & -- & -- & 5.90 & 157.8 & 1.42 & 305.7 \\
TexTok-256~\citep{zha2024language} & 675M & KL & 176M & 256 & 0.69 & -- & -- & 1.46 & 303.1 \\
LightningDiT~\citep{yao2025reconstruction} & 675M & KL & 70M & 256 & 0.28 & 2.17 & 205.6 & 1.35 & 295.3 \\

\rowcolor{gray!10}
MAETok + LightningDiT & 675M & AE & 176M & 128 & 0.48 & 2.21 & 208.3 & 1.73 & 308.4 \\
\rowcolor{gray!10}
MAETok + SiT-XL & 675M & AE & 176M & 128 & 0.48 & 2.31 & 216.5 & 1.67 & 311.2 \\
\midrule

\multicolumn{10}{l}{\textit{Ours}} \\
SMAP(VQ) + LlamaGen~\citep{sun2024llama} & 3.1B & VQ & 185M & 128 & 1.47 & 2.86 & 233.7 & 2.14 & 290.5 \\
\rowcolor{gray!10}
SMAP(KL) + CARD & 568M & KL & 391M & 128 & 0.75 & 2.38 & 244.6 & 1.97 & 320.8 \\
\rowcolor{gray!10}
SMAP(KL) + CARD & 1.1B & KL & 391M & 128 & 0.75 & 2.34 & 251.4 & 1.85 & 325.1 \\
\rowcolor{gray!10}
SMAP(SoftVQ) + CARD & 568M & SoftVQ & 391M & 128 & 0.55 & 2.69 & 211.3 & 2.01 & 304.8 \\
\rowcolor{gray!10}
SMAP(SoftVQ) + CARD & 1.1B & SoftVQ & 391M & 128 & 0.55 & 2.28 & 245.3 & 1.79 & 328.9 \\
\bottomrule
\end{tabular}
}
\vspace{-4mm}
\end{table*}

\noindent\textbf{Semantic Understanding through Conditional Embedding Injection.}
We perform a multi-stage analysis to verify that conditional embedding injection is not merely incidental, but instead plays a functional role in both tokenizer learning and downstream generation.

We first examine whether the injected semantic condition can itself serve as a meaningful source of global structure. As shown in~\autoref{fig:teaser}, when only the class condition is provided to the decoder, \texttt{SMAP} is already able to reconstruct coarse images that capture recognizable category-level semantics and rough global layout. Although these reconstructions are still blurry and lack instance-specific details, they are far from arbitrary outputs: the reconstructed content already reflects the semantic commonalities associated with the conditioning signal. This indicates that the semantic embedding is not treated merely as side information, but is explicitly trained to carry information that is directly useful for reconstruction.
Besides, we study how semantic information interacts with latent tokens once additional latent capacity is introduced. Again in~\autoref{fig:teaser}, adding latent tokens on top of the class condition leads to consistent improvements in reconstruction fidelity. This behavior reveals a clear division of roles: the conditional embedding establishes category-level identity and coarse global structure, while latent tokens progressively recover finer instance-level appearance, texture, and spatial details. Importantly, the semantic signal remains effective throughout this process rather than being overridden as more latent tokens are introduced, suggesting that semantic information is preserved as a stable prefix-level component of the learned representation.

To further probe this role decomposition, we visualize cross reconstructions in~\autoref{fig:ablation_cross}, where the semantic source $C$ and latent tokens $Z$ are independently manipulated across two input images. When only the semantic condition is retained ($Z=\emptyset$), the model still produces semantically recognizable reconstructions, again confirming that semantic identity can be recovered from the learned conditional prefix alone. When $C$ from one image is combined with $Z$ from another, the resulting reconstruction follows the semantic identity specified by $C$ while inheriting instance-level appearance cues from $Z$. In other words, the semantic condition primarily determines category-level identity, whereas latent tokens contribute instance-specific visual details. This provides direct evidence that \texttt{SMAP} learns a semantically grounded representation in which semantic prefixes and latent tokens play complementary and clearly differentiated roles.

Taken together, these findings show that conditional embedding injection does more than provide weak semantic alignment. Instead, it realizes semantic-aware prefix learning: semantic conditions are forced to encode category-level identity and global structure, while latent tokens progressively refine the representation with instance-level detail, yielding a latent space that benefits reconstruction.

\begin{figure*}[ht]
  \vskip 0.2in
  \begin{center}
    \centerline{\includegraphics[width=0.6 \textwidth]{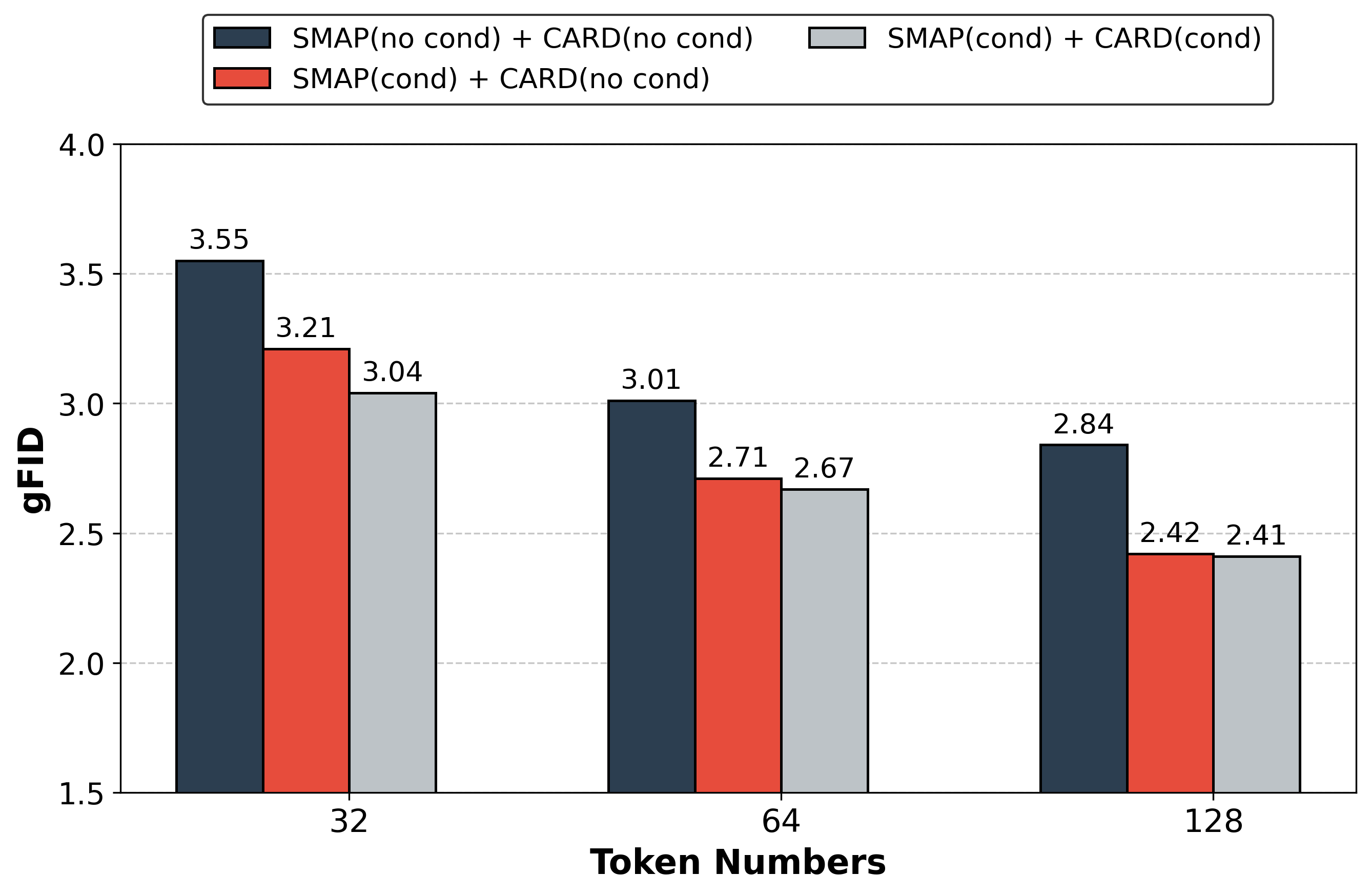}}
    \caption{\textbf{Effect of semantic-aware tokenization on downstream generation.}
    We compare three settings: a reconstruction-only tokenizer with independent generator conditioning, a semantic-aware tokenizer with independent generator conditioning, and a shared-semantic setting in which the generator reuses the tokenizer's learned semantic embedding space.
    Semantic-aware tokenizer pretraining consistently improves $\gFID$ across all token budgets, and semantic sharing yields a further gain in every setting.}
    \label{fig:ablation-tokenforgen}
  \end{center}
  \vspace{-4mm}
\end{figure*}

\noindent\textbf{Enforcing Semantic Dependency with Progressive Token Truncation.}
The emergence of semantically meaningful representations in \texttt{SMAP} is not solely a consequence of architectural design, but is critically driven by the proposed progressive token truncation strategy.
By truncating the suffix of the latent token sequence during training, the model is forced to shift increasing reconstruction responsibility toward the semantic condition and the early latent prefix.
As a result, the decoder can no longer rely exclusively on the full latent token set, and must instead learn to recover category-level identity and coarse global structure from the conditional prefix itself.
This behavior is clearly illustrated in~\autoref{fig:teaser}.
When only the class condition is provided, the model already produces coarse yet semantically recognizable reconstructions, indicating that the semantic prefix has absorbed meaningful global structural information.
When latent tokens are added back, reconstruction fidelity improves substantially and instance-specific details progressively reappear.
This shows that progressive token truncation does not simply act as a generic regularizer; rather, it explicitly encourages an information-ordered representation in which semantic conditions provide the structural scaffold and latent tokens refine it with finer visual detail.

The quantitative results in~\autoref{tab:ablation-generation-table} support the same conclusion.
Across all VQ settings, progressive token truncation improves both $\rFID$ and IS, with especially clear gains under limited token budgets. These qualitative and quantitative results show that progressive token truncation is the key mechanism that makes semantic information indispensable during training.
Without it, semantic embeddings are much less likely to develop into a functional reconstruction pathway; with it, they are explicitly forced to encode category-level commonality and global structure, thereby realizing semantic-aware prefix learning rather than merely adding auxiliary semantic supervision.

\subsection{Effect of Semantic-Aware Tokenization on Generation}

\noindent\textbf{Semantic-aware tokenization improves downstream generation.}
We next examine whether the semantic structure learned during tokenizer pretraining carries over to downstream generation. To isolate this effect, we train \texttt{CARD} on latent sequences produced by \texttt{SMAP} under three settings: (i) a reconstruction-only tokenizer with an independent generator-side conditioning pathway, (ii) a semantic-aware tokenizer while keeping the generator conditioning independent, and (iii) a shared-semantic setting in which the generator reuses the semantic embedding space learned during tokenizer pretraining.

As shown in~\autoref{fig:ablation-tokenforgen}, the benefit of semantic-aware tokenization is already evident even without semantic sharing on the generator side. Replacing the reconstruction-only tokenizer with the semantic-aware variant consistently improves $\gFID$ across all token budgets. This trend suggests that the gain is not solely due to a stronger conditioning mechanism during generation. Instead, semantic-aware tokenizer pretraining itself yields a latent space that is easier for the generator to model.
Reusing the tokenizer's learned semantic embedding space in the generator brings a further, though smaller, improvement. 
While these additional gains are more modest than those obtained from semantic-aware tokenization itself, their consistency indicates that the semantic representation learned by \texttt{SMAP} is not only transferable, but also directly useful for downstream generation.

This effect is also reflected at the system level in~\autoref{tab:main_256}. Under a compact 128-token budget, \texttt{SMAP}+\texttt{CARD} achieves competitive conditional generation performance against strong autoregressive and diffusion-based baselines. In particular, the 1.1B \texttt{SMAP(KL)} + \texttt{CARD} model reaches $\gFID=1.85$ with CFG, while the 1.1B \texttt{SMAP(SoftVQ)} + \texttt{CARD} model further improves to $\gFID=1.79$. These results are obtained with substantially fewer latent tokens than many prior methods, including approaches based on 256-4096 tokens, suggesting that the semantic-aware latent structure learned by \texttt{SMAP} supports efficient and high-quality generation.

Taken together, the semantic-aware prefix learning tokenizer produces a more generator-friendly latent representation even when tokenization and generation are trained separately. Besides, the learned semantic space can be reused by the generator to further improve alignment between tokenizer pretraining and downstream synthesis. Overall, the benefit of semantic-aware tokenization is therefore not limited to reconstruction quality but it leads to a latent representation that transfers more effectively to generative modeling.

\vspace{-5pt}

\section{Conclusion}
In this paper, we presented \texttt{SMAP}, a semantic-aware tokenizer that makes semantic information a functional component of tokenizer pretraining rather than a weak alignment signal. Through conditional embedding injection and tail token dropping, \texttt{SMAP} learns semantically grounded, information-ordered latent representations that improve reconstruction quality across discrete and continuous tokenization settings. Building on this latent space, we further introduced \texttt{CARD}, a hybrid autoregressive--diffusion generator that leverages the learned semantic structure for conditional image synthesis. Experiments on ImageNet show that semantic-aware prefix learning benefits both tokenizer reconstruction and downstream generation, suggesting that semantics should be treated as an integral part of representation learning in latent image modeling.

\nocite{langley00}

\bibliography{example_paper}

@STRING{jan  = {January}}

@STRING{jun  = {June}}

@string{nips      = {NeurIPS}}

@string{neurips   = {NeurIPS}}

@string{iccv      = {ICCV}}

@string{eccv      = {ECCV}}

@string{cvpr      = {CVPR}}

@string{icml      = {ICML}}

@string{iclr      = {ICLR}}

@string{aaai      = {Proceedings of the Conference on Articial Intelligence}}

@string{ijcv      = {International journal of Computer Vision}}

@string(arxiv   = {arXiv})

@inproceedings{langley00,
 author    = {P. Langley},
 title     = {Crafting Papers on Machine Learning},
 year      = {2000},
 pages     = {1207--1216},
 editor    = {Pat Langley},
 booktitle     = icml,
 address   = {Stanford, CA},
 publisher = {Morgan Kaufmann}
}

@inproceedings{esser2021taming,
  title={Taming transformers for high-resolution image synthesis},
  author={Esser, Patrick and Rombach, Robin and Ommer, Bjorn},
  booktitle=cvpr,
  pages={4195--4205},
  year={2021}
}

@inproceedings{rombach2022high,
  title={High-resolution image synthesis with latent diffusion models},
  author={Rombach, Robin and Blattmann, Andreas and Lorenz, Dominik and Esser, Patrick and Ommer, Bj{\"o}rn},
  booktitle=cvpr,
  pages={10684--10695},
  year={2022}
}

@inproceedings{ma2024sit, 
title={SiT: Exploring Flow and Diffusion-Based Generative Models with Scalable Interpolant Transformers}, 
author={Ma, Nanye and Goldstein, Mark and Albergo, Michael S and Boffi, Nicholas M and Vanden-Eijnden, Eric and Xie, Saining}, 
booktitle=eccv,
pages={23–40},
year={2024}}

@inproceedings{peebles2023dit,
  title={Scalable diffusion models with transformers},
  author={Peebles, William and Xie, Saining},
  booktitle={Proceedings of the IEEE/CVF International Conference on Computer Vision},
  pages={4195--4205},
  year={2023}
}

@article{qiu2025robust,
  title={Robust Latent Matters: Boosting Image Generation with Sampling Error Synthesis},
  author={Qiu, Kai and Li, Xiang and Kuen, Jason and Chen, Hao and Xu, Xiaohao and Gu, Jiuxiang and Luo, Yinyi and Raj, Bhiksha and Lin, Zhe and Savvides, Marios},
  journal={arXiv preprint arXiv:2503.08354},
  year={2025}
}

@article{zha2024language,
  title={Language-Guided Image Tokenization for Generation},
  author={Zha, Kaiwen and Yu, Lijun and Fathi, Alireza and Ross, David A and Schmid, Cordelia and Katabi, Dina and Gu, Xiuye},
  journal={arXiv preprint arXiv:2412.05796},
  year={2024}
}

@article{yao2025reconstruction,
  title={Reconstruction vs. Generation: Taming Optimization Dilemma in Latent Diffusion Models},
  author={Yao, Jingfeng and Wang, Xinggang},
  journal={arXiv preprint arXiv:2501.01423},
  year={2025}
}

@InProceedings{kingma2014, 
title={Auto-Encoding Variational Bayes},
booktitle = iclr, 
author={Kingma, Diederik P and Welling, Max}, 
year={2014}}

@inproceedings{
yao2024fasterdit,
title={FasterDiT: Towards Faster Diffusion Transformers Training without Architecture Modification},
author={Jingfeng Yao and Cheng Wang and Wenyu Liu and Xinggang Wang},
booktitle=neurips,
year={2024},
}

@inproceedings{oord2017vq,
 author = {van den Oord, Aaron and Vinyals, Oriol and kavukcuoglu, koray},
 booktitle = nips,
 title = {Neural Discrete Representation Learning},
 volume = {30},
 year = {2017}
}

@inproceedings{
Yu2025repa,
title={Representation Alignment for Generation: Training Diffusion Transformers Is Easier Than You Think},
author={Sihyun Yu and Sangkyung Kwak and Huiwon Jang and Jongheon Jeong and Jonathan Huang and Jinwoo Shin and Saining Xie},
booktitle=iclr,
year={2025},
}

@INPROCEEDINGS{deng2009imagenet,
  author={Deng, Jia and Dong, Wei and Socher, Richard and Li, Li-Jia and Kai Li and Li Fei-Fei},
  booktitle=cvpr, 
  title={ImageNet: A large-scale hierarchical image database}, 
  year={2009},
  pages={248-255},}

@inproceedings{
dhariwal2021adm,
title={Diffusion Models Beat {GAN}s on Image Synthesis},
author={Prafulla Dhariwal and Alexander Quinn Nichol},
booktitle=nips,
year={2021},
}

@inproceedings{li2023mage,
  title={Mage: Masked generative encoder to unify representation learning and image synthesis},
  author={Li, Tianhong and Chang, Huiwen and Mishra, Shlok and Zhang, Han and Katabi, Dina and Krishnan, Dilip},
  booktitle=cvpr,
  pages={2142--2152},
  year={2023}
}

@article{li2024mar,
  title={Autoregressive Image Generation without Vector Quantization},
  author={Li, Tianhong and Tian, Yonglong and Li, He and Deng, Mingyang and He, Kaiming},
  journal={arXiv preprint arXiv:2406.11838},
  year={2024}
}

@article{tian2024var,
  title={Visual autoregressive modeling: Scalable image generation via next-scale prediction},
  author={Tian, Keyu and Jiang, Yi and Yuan, Zehuan and Peng, Bingyue and Wang, Liwei},
  journal={arXiv preprint arXiv:2404.02905},
  year={2024}
}

@misc{nichol2021ddpm,
      title={Improved Denoising Diffusion Probabilistic Models}, 
      author={Alex Nichol and Prafulla Dhariwal},
      year={2021},
      eprint={2102.09672},
      archivePrefix={arXiv},
      primaryClass={cs.LG},
      url={https://arxiv.org/abs/2102.09672}, 
}

@article{dhariwal2021diffusion,
  title={Diffusion models beat gans on image synthesis},
  author={Dhariwal, Prafulla and Nichol, Alexander},
  journal={Advances in neural information processing systems},
  volume={34},
  pages={8780--8794},
  year={2021}
}

@misc{song2022ddim,
      title={Denoising Diffusion Implicit Models}, 
      author={Jiaming Song and Chenlin Meng and Stefano Ermon},
      year={2022},
      eprint={2010.02502},
      archivePrefix={arXiv},
      primaryClass={cs.LG},
      url={https://arxiv.org/abs/2010.02502}, 
}

@misc{vahdat2021scorelatentdiffusion,
      title={Score-based Generative Modeling in Latent Space}, 
      author={Arash Vahdat and Karsten Kreis and Jan Kautz},
      year={2021},
      eprint={2106.05931},
      archivePrefix={arXiv},
      primaryClass={stat.ML},
      url={https://arxiv.org/abs/2106.05931}, 
}

@misc{rombach2022ldm,
      title={High-Resolution Image Synthesis with Latent Diffusion Models}, 
      author={Robin Rombach and Andreas Blattmann and Dominik Lorenz and Patrick Esser and Björn Ommer},
      year={2022},
      eprint={2112.10752},
      archivePrefix={arXiv},
      primaryClass={cs.CV},
      url={https://arxiv.org/abs/2112.10752}, 
}

@misc{sohldickstein2015deepunsupervisedlearningusing,
      title={Deep Unsupervised Learning using Nonequilibrium Thermodynamics}, 
      author={Jascha Sohl-Dickstein and Eric A. Weiss and Niru Maheswaranathan and Surya Ganguli},
      year={2015},
      eprint={1503.03585},
      archivePrefix={arXiv},
      primaryClass={cs.LG},
      url={https://arxiv.org/abs/1503.03585}, 
}

@inproceedings{tschannen2025givt,
  title={Givt: Generative infinite-vocabulary transformers},
  author={Tschannen, Michael and Eastwood, Cian and Mentzer, Fabian},
  booktitle=eccv,
  pages={292--309},
  year={2025},
}

@article{zhu2023designing,
  title={Designing a better asymmetric vqgan for stablediffusion},
  author={Zhu, Zixin and Feng, Xuelu and Chen, Dongdong and Bao, Jianmin and Wang, Le and Chen, Yinpeng and Yuan, Lu and Hua, Gang},
  journal={arXiv preprint arXiv:2306.04632},
  year={2023}
}

@inproceedings{takahashi2019variational,
  title={Variational autoencoder with implicit optimal priors},
  author={Takahashi, Hiroshi and Iwata, Tomoharu and Yamanaka, Yuki and Yamada, Masanori and Yagi, Satoshi},
  booktitle={Proceedings of the AAAI Conference on Artificial Intelligence},
  volume={33},
  pages={5066--5073},
  year={2019}
}

@inproceedings{
    sahoo2024mdtv2,
    title={Simple and Effective Masked Diffusion Language Models},
    author={Subham Sekhar Sahoo and Marianne Arriola and Aaron Gokaslan and Edgar Mariano Marroquin and Alexander M Rush and Yair Schiff and Justin T Chiu and Volodymyr Kuleshov},
    booktitle={The Thirty-eighth Annual Conference on Neural Information Processing Systems},
    year={2024},
    url={https://openreview.net/forum?id=L4uaAR4ArM}
}

@article{van2017neural,
  title={Neural discrete representation learning},
  author={Van Den Oord, Aaron and Vinyals, Oriol and others},
  journal={Advances in neural information processing systems},
  volume={30},
  year={2017}
}

@article{razavi2019generating,
  title={Generating diverse high-fidelity images with vq-vae-2},
  author={Razavi, Ali and Van den Oord, Aaron and Vinyals, Oriol},
  journal={Advances in neural information processing systems},
  volume={32},
  year={2019}
}

@article{yu2021vector,
  title={Vector-quantized image modeling with improved vqgan},
  author={Yu, Jiahui and Li, Xin and Koh, Jing Yu and Zhang, Han and Pang, Ruoming and Qin, James and Ku, Alexander and Xu, Yuanzhong and Baldridge, Jason and Wu, Yonghui},
  journal={arXiv preprint arXiv:2110.04627},
  year={2021}
}

@inproceedings{zheng2023online,
  title={Online clustered codebook},
  author={Zheng, Chuanxia and Vedaldi, Andrea},
  booktitle={ICCV},
  year={2023}
}

@article{hinton2006reducing,
  title={Reducing the dimensionality of data with neural networks},
  author={Hinton, Geoffrey E and Salakhutdinov, Ruslan R},
  journal={science},
  volume={313},
  number={5786},
  pages={504--507},
  year={2006},
  publisher={American Association for the Advancement of Science}
}

@inproceedings{vincent2008extracting,
  title={Extracting and composing robust features with denoising autoencoders},
  author={Vincent, Pascal and Larochelle, Hugo and Bengio, Yoshua and Manzagol, Pierre-Antoine},
  booktitle={Proceedings of the 25th international conference on Machine learning},
  pages={1096--1103},
  year={2008}
}

@inproceedings{
yu2022vectorquantized,
title={Vector-quantized Image Modeling with Improved {VQGAN}},
author={Jiahui Yu and Xin Li and Jing Yu Koh and Han Zhang and Ruoming Pang and James Qin and Alexander Ku and Yuanzhong Xu and Jason Baldridge and Yonghui Wu},
booktitle=iclr,
year={2022},}

@inproceedings{
yu2024language,
title={Language Model Beats Diffusion - Tokenizer is key to visual generation},
author={Lijun Yu and Jose Lezama and Nitesh Bharadwaj Gundavarapu and Luca Versari and Kihyuk Sohn and David Minnen and Yong Cheng and Agrim Gupta and Xiuye Gu and Alexander G Hauptmann and Boqing Gong and Ming-Hsuan Yang and Irfan Essa and David A Ross and Lu Jiang},
booktitle={The Twelfth International Conference on Learning Representations},
year={2024},
}

@article{gpt3,
  title={Language Models are Few-Shot Learners},
  author={Brown, Tom B. and Mann, Benjamin and Ryder, Nick and Subbiah, Melanie and Kaplan, Jared and Dhariwal, Prafulla and Neelakantan, Arvind and Shyam, Pranav and Sastry, Girish and Askell, Amanda and others},
  journal=NIPS,
  year={2020},
  url={https://proceedings.neurips.cc/paper/2020/file/1457c0d6bfcb4967418bfb8ac142f64a-Paper.pdf},
}

@article{loshchilov2017decoupled,
  title={Decoupled weight decay regularization},
  author={Loshchilov, Ilya and Hutter, Frank},
  journal={arXiv preprint arXiv:1711.05101},
  year={2017}
}

@article{liu2024customize,
  title={Customize your visual autoregressive recipe with set autoregressive modeling},
  author={Liu, Wenze and Zhuo, Le and Xin, Yi and Xia, Sheng and Gao, Peng and Yue, Xiangyu},
  journal={arXiv preprint arXiv:2410.10511},
  year={2024}
}

@inproceedings{lee2022rqvae,
  title={Autoregressive image generation using residual quantization},
  author={Lee, Doyup and Kim, Chiheon and Kim, Saehoon and Cho, Minsu and Han, Wook-Shin},
  booktitle={Proceedings of the IEEE/CVF Conference on Computer Vision and Pattern Recognition},
  pages={11523--11532},
  year={2022}
}

@article{vaswani2017attention,
  title={Attention is all you need},
  author={Vaswani, Ashish and Shazeer, Noam and Parmar, Niki and Uszkoreit, Jakob and Jones, Llion and Gomez, Aidan N and Kaiser, {\L}ukasz and Polosukhin, Illia},
  journal=NIPS,
  year={2017}
}

@inproceedings{yu2025randomized,
  title={Randomized Autoregressive Visual Generation},
  author={Yu, Qihang and He, Ju and Deng, Xueqing and Shen, Xiaohui and Chen, Liang-Chieh},
  booktitle={ICCV},
  year={2025}
}

@inproceedings{chang2022maskgit,
  title={Maskgit: Masked generative image transformer},
  author={Chang, Huiwen and Zhang, Han and Jiang, Lu and Liu, Ce and Freeman, William T},
  booktitle={CVPR},
  year={2022}
}

@inproceedings{kingma2013vae,
  title={Auto-encoding variational bayes},
  author={Kingma, Diederik P and Welling, Max},
  booktitle=ICLR,
  year={2014}
}

@article{van2016conditional,
  title={Conditional image generation with pixelcnn decoders},
  author={Van den Oord, Aaron and Kalchbrenner, Nal and Espeholt, Lasse and Vinyals, Oriol and Graves, Alex and others},
  journal=NIPS,
  year={2016}
}

@inproceedings{yu2024titok,
title={An image is worth 32 tokens for reconstruction and generation},
author={Qihang Yu and Mark Weber and Xueqing Deng and Xiaohui Shen and Daniel Cremers and Liang-Chieh Chen},
booktitle={The Thirty-eighth Annual Conference on Neural Information Processing Systems},
year={2024}
}

@inproceedings{kim2025tatitok,
title={Democratizing text-to-image masked generative models with compact text-aware one-dimensional tokens},
author={Dongwon Kim and Ju He and Qihang Yu and Chenglin Yang and Xiaohui Shen and Suha Kwak and Liang-Chieh Chen},
booktitle=iccv,
year={2025}
}

@inproceedings{miwa2025onedpiece,
title= {One-D-Piece: Image Tokenizer Meets Quality-Controllable Compression},
author= {Keita Miwa and Kento Sasaki and Hidehisa Arai and Tsubasa Takahashi and Yu Yamaguchi},
booktitle={arXiv preprint arXiv:2501.10064},
year={2025}
}

@inproceedings{li2024imagefolder,
title={ImageFolder: Autoregressive Image Generation with Folded Tokens},
author={Li, Xiang and Chen, Hao and Qiu, Kai and Kuen, Jason and Gu, Jiuxiang and Raj, Bhiksha and Lin, Zhe},
booktitle={arXiv preprint arXiv:2410.01756},
year={2024}
}

@inproceedings{Radford2021clip,
title={Learning Transferable Visual Models From Natural Language Supervision},
author={Alec Radford and Jong Wook Kim and Chris Hallacy and Aditya Ramesh and Gabriel Goh and Sandhini Agarwal and Girish Sastry and Amanda Askell and Pamela Mishkin and Jack Clark and Gretchen Krueger and Ilya Sutskever 1},
booktitle=PmLR,
year={2021}
}

@inproceedings{li2023blip2,
title={{BLIP-2:} Bootstrapping Language-Image Pre-training with Frozen Image Encoders and Large Language Models}, 
author={Junnan Li and Dongxu Li and Silvio Savarese and Steven Hoi},
year={2023},
booktitle={ICML},
}

@inproceedings{lipman2023flow,
title={Flow Matching for Generative Modeling},
author={Yaron Lipman and Ricky T. Q. Chen and Heli Ben-Hamu and Maximilian Nickel1 Matt Le},
booktitle=iclr,
year={2023}
}

@inproceedings{yao2025vtp,
title={Towards Scalable Pre-training of Visual Tokenizers for Generation},
author={Yao Jingfeng and Song Yuda and Zhou Yucong and Wang Xinggang},
booktitle={arXiv preprint arXiv:2512.13687},
year={2025}
}

@inproceedings{nextstepteam2025nextstep1,
title={NextStep-1: Toward Autoregressive Image Generation with Continuous Tokens at Scale},
author={NextStep Team and Chunrui Han and Guopeng Li and Jingwei Wu and Quan Sun and Yan Cai and Yuang Peng and Zheng Ge and Deyu Zhou and Haomiao Tang and Hongyu Zhou and Kenkun Liu and Ailin Huang and Bin Wang and Changxin Miao and Deshan Sun and En Yu and Fukun Yin and Gang Yu and Hao Nie and Haoran Lv and Hanpeng Hu and Jia Wang and Jian Zhou and Jianjian Sun and Kaijun Tan and Kang An and Kangheng Lin and Liang Zhao and Mei Chen and Peng Xing and Rui Wang and Shiyu Liu and Shutao Xia and Tianhao You and Wei Ji and Xianfang Zeng and Xin Han and Xuelin Zhang and Yana Wei and Yanming Xu and Yimin Jiang and Yingming Wang and Yu Zhou and Yucheng Han and Ziyang Meng and Binxing Jiao and Daxin Jiang and Xiangyu Zhang and Yibo Zhu},
booktitle={arXiv preprint arXiv:2508.10711},
year={2025}
}

@inproceedings{ke2025hyperspherical,
title={Hyperspherical Latents Improve Continuous-Token Autoregressive Generation}, 
author={Guolin Ke and Hui Xue},
booktitle={arXiv preprint arXiv:2509.24335},
year={2025}
}

@inproceedings{chen2025softvqvae,
title={SoftVQ-VAE: Efficient 1-Dimensional Continuous Tokenizer},
author={Hao Chen and Ze Wang and Xiang Li and Ximeng Sun and Fangyi Chen and Jiang Liu and Jindong Wang and Bhiksha Raj and Zicheng Liu and Emad Barsoum},
booktitle={CVPR},
year={2025},
}

@inproceedings{sun2024llama,
  title={Autoregressive Model Beats Diffusion: Llama for Scalable Image Generation},
  author={Sun, Peize and Jiang, Yi and Chen, Shoufa and Zhang, Shilong and Peng, Bingyue and Luo, Ping and Yuan, Zehuan},
  booktitle={arXiv preprint arXiv:2406.06525},
  year={2024}
}

@article{kuznetsova2020open,
  title={The open images dataset v4: Unified image classification, object detection, and visual relationship detection at scale},
  author={Kuznetsova, Alina and Rom, Hassan and Alldrin, Neil and Uijlings, Jasper and Krasin, Ivan and Pont-Tuset, Jordi and Kamali, Shahab and Popov, Stefan and Malloci, Matteo and Kolesnikov, Alexander and others},
  journal=IJCV,
  year={2020},
}

@inproceedings{schuhmann2022laion,
  title={Laion-5b: An open large-scale dataset for training next generation image-text models},
  author={Schuhmann, Christoph and Beaumont, Romain and Vencu, Richard and Gordon, Cade and Wightman, Ross and Cherti, Mehdi and Coombes, Theo and Katta, Aarush and Mullis, Clayton and Wortsman, Mitchell and others},
  booktitle={The Thirty-sixth Annual Conference on Neural Information Processing Systems},
  year={2022}
}

@article{fid,
  title={Gans trained by a two time-scale update rule converge to a local nash equilibrium},
  author={Heusel, Martin and Ramsauer, Hubert and Unterthiner, Thomas and Nessler, Bernhard and Hochreiter, Sepp},
  journal={Advances in neural information processing systems},
  volume={30},
  year={2017}
}

@article{is,
  title={Improved techniques for training gans},
  author={Salimans, Tim and Goodfellow, Ian and Zaremba, Wojciech and Cheung, Vicki and Radford, Alec and Chen, Xi},
  journal={Advances in neural information processing systems},
  volume={29},
  year={2016}
}

@inproceedings{bao2023all,
  title={All are worth words: A vit backbone for diffusion models},
  author={Bao, Fan and Nie, Shen and Xue, Kaiwen and Cao, Yue and Li, Chongxuan and Su, Hang and Zhu, Jun},
  booktitle=CVPR,
  year={2023}
}

@article{yu2022scaling,
  title={Scaling autoregressive models for content-rich text-to-image generation},
  author={Yu, Jiahui and Xu, Yuanzhong and Koh, Jing Yu and Luong, Thang and Baid, Gunjan and Wang, Zirui and Vasudevan, Vijay and Ku, Alexander and Yang, Yinfei and Ayan, Burcu Karagol and others},
  journal={arXiv preprint arXiv:2206.10789},
  year={2022}
}

@article{goodfellow2014generative,
  title={Generative adversarial nets},
  author={Goodfellow, Ian and Pouget-Abadie, Jean and Mirza, Mehdi and Xu, Bing and Warde-Farley, David and Ozair, Sherjil and Courville, Aaron and Bengio, Yoshua},
  journal={Advances in neural information processing systems},
  volume={27},
  year={2014}
}

@misc{ryu2024vqgan,
  author       = {Simo Ryu},
  title        = {Training VQGAN and VAE, with Detailed Explanation},
  year         = {2024},
  howpublished = {\url{https://github.com/cloneofsimo/vqgan-training}},
  note         = {GitHub repository}
}
\bibliographystyle{icml2026}

\newpage
\appendix
\onecolumn

\appendix
\section*{Appendix}
\label{sec:supplementary}

In the supplementary materials, we provide the following additional details:
\vspace{-3mm}
\begin{itemize}
    \item The comprehensive training and testing hyper-parameters and training costs for \modelname (\secref{app:implementation_details}).
    \vspace{-2mm}
    \item The detailed instantiation of the unified regularization operator $\mathtt{Regu}(\cdot)$ under both discrete and continuous tokenization settings (\secref{app:regu}).
    \vspace{-2mm}
    \item A more comprehensive comparison with more metrics and baselines (\secref{app:detailed_preliminary}).
    \vspace{-2mm}
    \item Limitation discussion (\secref{app:limitations}).
    \vspace{-2mm}
    \item Dataset Licenses (\secref{app:dataset_license}).
\end{itemize}

\section{Additional Implementation Details}
\label{app:implementation_details}

\noindent \textbf{Tokenizer training.}
For image reconstruction (\textbf{tokenizer}), we train \modelname\ on ImageNet following the one-stage recipe described in the main paper. Unless otherwise specified, the training augmentation is limited to random resized cropping and horizontal flipping. All tokenizer variants are trained for \(500\mathrm{K}\) iterations at resolutions \(256\times256\) and \(512\times512\), respectively. We use the AdamW optimizer~\cite{loshchilov2017decoupled} with batch size \texttt{256}, initial learning rate \texttt{\(1\times 10^{-4}\)}, and weight decay \texttt{\(1\times 10^{-4}\)}. The learning rate follows a cosine decay schedule with \texttt{20} warm-up epochs. We use patch size \(16\) for all Vision Transformer tokenizers at resolution \(256\times256\), and increase it to \(32\) at resolution \(512\times512\) for better computational efficiency.

\modelname-S, \modelname-B, and \modelname-L denote the small, base, and large tokenizer variants, with parameter counts of \(185\)M, \(391\)M, and \(568\)M, respectively. For the \texttt{VQ} variant, we use a codebook of size \(8192\) with code dimension \(64\), and train models with token lengths \(64\) and \(128\). For the \texttt{KL} variant, following~\cite{li2024mar}, we use continuous latent features with \(16\) channels and token lengths \(128\) and \(256\). For the \texttt{SoftVQ} variant, we adopt a four-level hierarchical codebook with total size \(8192\), while keeping the channel dimension at \(16\); the token lengths are also set to \(128\) and \(256\). During training, the retained prefix length \(k\) is sampled uniformly from \(\{0,1,\dots,K\}\), such that the decoder is exposed to variable token budgets throughout optimization. The class embedding module is trained jointly with the tokenizer, and the same semantic condition embedding is used in both the encoder and decoder.

\noindent \textbf{Generator training.}
For image generation (\textbf{generator}), we use different generators for discrete and continuous latent spaces. For discrete latent tokenization, we adopt \texttt{LlamaGen}~\cite{sun2024llama} following the standard setup for class-conditional generation. For continuous latent tokenization, we train the proposed \CARD\ model on top of the latent space produced by \modelname. \CARD-B, \CARD-L, and \CARD-XL contain \(234\)M, \(568\)M, and \(1.1\)B parameters, respectively, and their detailed architectural configurations are reported in \autoref{tab:CARD_configuration}. Unless otherwise specified, the generator is trained with batch size \texttt{2048} for \texttt{250k} iterations using AdamW~\cite{loshchilov2017decoupled}, with learning rate \texttt{\(2\times 10^{-4}\)} and weight decay \texttt{\(1\times 10^{-5}\)}. We use cosine learning-rate decay and apply class-condition dropout with probability \texttt{0.1} for classifier-free guidance.

For \CARD, the class condition is not produced by a separate encoder. Instead, we directly reuse the semantic embedding module learned during \modelname\ pretraining, which ensures semantic consistency between tokenization and generation. During evaluation, we follow prior work~\cite{dhariwal2021adm,esser2021taming} and report \(\gFID\) and IS over \(50{,}000\) generated samples. For class-conditional sampling, we use classifier-free guidance with guidance scale \texttt{[2.7 for 256]} at \(256\times256\) and \texttt{[3.5 for 512]} at \(512\times512\). For the discrete \texttt{LlamaGen} results, we use the decoding and sampling hyper-parameters from the official implementation unless otherwise specified. For the continuous \CARD\ results, we use \texttt{25} flow-matching sampling steps as the default inference configuration.

\noindent \textbf{Training cost.}
The tokenizer training takes \texttt{32} A800 GPUs for \texttt{32} hours for \modelname-S, \texttt{32} A800 GPUs for \texttt{32} hours for \modelname-B, and \texttt{32} A800 GPUs for \texttt{72} hours for \modelname-L. The generator training takes \texttt{16} A800 GPUs for \texttt{48} hours for \CARD-B, \texttt{32} A800 GPUs for \texttt{48} hours for \CARD-L, and \texttt{32} A800 GPUs for \texttt{96} hours for \CARD-XL.

\section{Instantiation of the Unified Regularization Operator \texorpdfstring{$\mathtt{Regu}(\cdot)$}{Regu(.)}}
\label{app:regu}

In the main text, we use $\mathtt{Regu}(\cdot)$ as a unified notation for the latent regularization applied to the 1D latent tokens before decoding. This abstraction allows us to describe discrete and continuous tokenizers within a single encoder--decoder formulation. In practice, however, as shown in~\autoref{tab:regu_instantiation}, $\mathtt{Regu}(\cdot)$ corresponds to different operations depending on the tokenizer instantiation.

Let $\mathbf{Z}_{\mathrm{1D}} = [\mathbf{z}_1,\dots,\mathbf{z}_K] \in \mathbb{R}^{K \times D}$ denote the latent sequence produced by the encoder.

\paragraph{VQ instantiation.}
For the discrete VQ tokenizer, $\mathtt{Regu}(\cdot)$ denotes vector quantization with a learned codebook $\mathcal{E} = \{\mathbf{e}_1,\dots,\mathbf{e}_{|\mathcal{E}|}\}$.
Each latent token $\mathbf{z}_i$ is replaced by its nearest codebook entry:
\begin{equation}
\mathtt{Regu}(\mathbf{z}_i)
=
\mathbf{e}_{j^\star},
\qquad
j^\star
=
\arg\min_j \|\mathbf{z}_i - \mathbf{e}_j\|_2^2.
\end{equation}
Applying this token-wise operation to the full sequence yields the quantized latent sequence
\begin{equation}
\mathtt{Regu}(\mathbf{Z}_{\mathrm{1D}})
=
[\mathbf{e}_{j_1^\star}, \dots, \mathbf{e}_{j_K^\star}].
\end{equation}
This is the discrete latent representation used by the decoder.

\paragraph{KL instantiation.}
For the continuous VAE-style tokenizer, $\mathtt{Regu}(\cdot)$ denotes variational regularization via Gaussian reparameterization.
The encoder predicts mean and variance parameters $(\boldsymbol{\mu}_i, \boldsymbol{\sigma}_i)$ for each latent token, and sampling is performed as
\begin{equation}
\mathtt{Regu}(\mathbf{z}_i)
=
\boldsymbol{\mu}_i + \boldsymbol{\sigma}_i \odot \boldsymbol{\epsilon}_i,
\qquad
\boldsymbol{\epsilon}_i \sim \mathcal{N}(\mathbf{0}, \mathbf{I}).
\end{equation}
Thus, for the KL case, $\mathtt{Regu}(\mathbf{Z}_{\mathrm{1D}})$ denotes the reparameterized continuous latent sequence passed to the decoder. During training, this is accompanied by the standard KL regularization term that encourages the posterior to remain close to a Gaussian prior.

\paragraph{SoftVQ instantiation.}
For the SoftVQ tokenizer, $\mathtt{Regu}(\cdot)$ denotes differentiable soft quantization rather than hard nearest-neighbor assignment.
Each latent token $\mathbf{z}_i$ is softly matched against the codebook entries, producing assignment weights
\begin{equation}
\alpha_{ij}
=
\frac{\exp(-d(\mathbf{z}_i,\mathbf{e}_j)/\tau)}
{\sum_{j'} \exp(-d(\mathbf{z}_i,\mathbf{e}_{j'})/\tau)},
\end{equation}
where $d(\cdot,\cdot)$ is a distance function and $\tau$ is a temperature parameter.
The regularized latent is then given by the weighted combination
\begin{equation}
\mathtt{Regu}(\mathbf{z}_i)
=
\sum_j \alpha_{ij}\mathbf{e}_j.
\end{equation}
Accordingly, $\mathtt{Regu}(\mathbf{Z}_{\mathrm{1D}})$ is a differentiably quantized latent sequence that retains codebook structure while remaining continuous during optimization.

\paragraph{Unifying view.}
Although these three instantiations differ algorithmically, they play the same functional role in our framework: they transform the encoder-produced latent sequence into the regularized representation consumed by the decoder. Using $\mathtt{Regu}(\cdot)$ therefore lets us present the SMAP tokenizer in a unified way while preserving the flexibility to instantiate it with discrete, continuous, or softly quantized latents.

\begin{table*}[t]
\caption{\textbf{Instantiation of $\mathtt{Regu}(\cdot)$ under different tokenizer formulations.}}
\label{tab:regu_instantiation}
\centering
\small
\setlength{\tabcolsep}{5pt}
\scalebox{0.8}{
\begin{tabular}{lcccc}
\toprule
Tokenizer & Encoder output & Regularization target & $\mathtt{Regu}(\cdot)$ instantiation & Output to decoder \\
\midrule
VQ
& $\mathbf{Z}_{\mathrm{1D}} \in \mathbb{R}^{K \times D}$
& discrete codebook
& nearest-neighbor vector quantization
& quantized code embeddings \\
KL
& $(\boldsymbol{\mu}, \boldsymbol{\sigma})$
& Gaussian posterior
& reparameterized latent sampling
& continuous sampled latents \\
SoftVQ
& $\mathbf{Z}_{\mathrm{1D}} \in \mathbb{R}^{K \times D}$
& soft code assignment
& temperature-controlled soft quantization
& weighted codebook mixtures \\
\bottomrule
\end{tabular}
}
\end{table*}

\section{Detailed Results of Preliminary Experiments}
\label{app:detailed_preliminary}
We summarize the detailed results of preliminary experiments in~\autoref{tab:detailed_results} and~\autoref{tab:imagenet_512_supp}.

\begin{table}[t]
\centering
\scriptsize
\caption{\textbf{Detailed results of preliminary experiments in the main paper.}}
\vspace{-2ex}
\label{tab:detailed_results}

\begin{subtable}{0.45\linewidth}
\centering
\caption{reconstruction FID (\texttt{VQ}).}
\label{tab:detailed_results_vq}
\begin{tabular}{l|cccccc}
\toprule
\#token & 32 & 64 & 96 & 128 & 192 & 256 \\
\midrule
\texttt{SMAP-VQ-S} & 4.012 & 3.297 & 2.711 & 2.217 & 1.706 & 1.524 \\
\texttt{SMAP-VQ-B} & 3.502 & 2.706 & 1.962 & 1.743 & 1.398 & 1.279 \\
\texttt{SMAP-VQ-L} & 3.241 & 2.191 & 1.687 & 1.479 & 1.175 & 1.012 \\
\bottomrule
\end{tabular}
\end{subtable}
\hfill
\begin{subtable}{0.45\linewidth}
\centering
\caption{reconstruction FID (\texttt{KL}).}
\label{tab:detailed_results_kl}
\begin{tabular}{l|cccccc}
\toprule
\#token & 32 & 64 & 96 & 128 & 192 & 256 \\
\midrule
\texttt{SMAP-KL-S} & 1.493 & 1.245 & 1.094 & 0.998 & 0.927 & 0.879 \\
\texttt{SMAP-KL-B} & 1.311 & 1.147 & 0.912 & 0.813 & 0.763 & 0.731 \\
\texttt{SMAP-KL-L} & 1.271 & 0.831 & 0.749 & 0.671 & 0.612 & 0.569 \\
\bottomrule
\end{tabular}
\end{subtable}

\vspace{1.2ex}

\begin{subtable}{0.45\linewidth}
\centering
\caption{reconstruction FID (\texttt{SoftVQ}).}
\label{tab:detailed_results_softvq}
\begin{tabular}{l|cccccc}
\toprule
\#token & 32 & 64 & 96 & 128 & 192 & 256 \\
\midrule
\texttt{SMAP-SoftVQ-S} & 1.320 & 0.997 & 0.918 & 0.879 & 0.755 & 0.713 \\
\texttt{SMAP-SoftVQ-B} & 1.287 & 0.862 & 0.751 & 0.694 & 0.611 & 0.575 \\
\texttt{SMAP-SoftVQ-L} & 1.191 & 0.701 & 0.502 & 0.420 & 0.389 & 0.366 \\
\bottomrule
\end{tabular}
\end{subtable}
\end{table}

\begin{table}
\centering
\small
\caption{\textbf{ImageNet-1K $512\times 512$ generation results evaluated with ADM~\cite{dhariwal2021diffusion}.} \dag: Trained on OpenImages~\cite{kuznetsova2020open} \ddag: Trained on OpenImages, LAION-Aesthetics/-Humans~\cite{schuhmann2022laion}. 
P: generator's parameters. S: sampling steps. T: throughput as samples per seconds on A100 with float32 precision, measured with \textit{w/ guidance} variants if available. ``guidance" refers to classifier-free guidance.
}
\label{tab:imagenet_512_supp}
\resizebox{0.75\textwidth}{!}{
\begin{tabular}{lc|cccccccc}
\toprule
\multirow{2}{*}{tokenizer} & \multirow{2}{*}{rFID$\downarrow$} & \multirow{2}{*}{generator} & \multicolumn{2}{c}{w/o guidance} & \multicolumn{2}{c}{w/ guidance} & \multirow{2}{*}{P$\downarrow$} & \multirow{2}{*}{S$\downarrow$} & \multirow{2}{*}{T$\uparrow$} \\
 &  &  & gFID$\downarrow$ & IS$\uparrow$ & gFID$\downarrow$ & IS$\uparrow$ & & & \\
\hline
\multicolumn{10}{c}{\textit{diffusion-based generative models}} \\
\hline
\multirow{3}{*}{VAE~\cite{}\ddag} & \multirow{3}{*}{0.19} & UViT-L/4~\cite{bao2023all} & 18.03 & 76.9 & 4.67 & 213.3 & 287M & 50 & 1.0 \\
 & & UViT-H/4~\cite{bao2023all} & 15.71 & 101.3 & 4.05 & 263.8 & 501M & 50 & 0.6 \\
 & & DiT-XL/2~\cite{peebles2023dit} & 12.03 & 105.3 & 3.04 & 240.8 & 675M & 250 & 0.1 \\
\hline
\multicolumn{10}{c}{\textit{transformer-based generative models}} \\
\hline
MaskGIT-VQGAN~\cite{chang2022maskgit} & 1.97 & MaskGIT-ViT~\cite{chang2022maskgit} & 7.32 & 156.0 & - & - & \textbf{177M} & 12 & 3.9 \\
\multirow{2}{*}{\textcolor{gray}{LFQ~\cite{yu2024language}}} & \multirow{2}{*}{\textcolor{gray}{1.22}} & \multirow{2}{*}{\textcolor{gray}{MAGVIT-v2~\cite{yu2024language}}} & \textcolor{gray}{4.61} & \textcolor{gray}{192.4} & - & - & \textcolor{gray}{307M} & \textcolor{gray}{12} & \textcolor{gray}{3.5} \\
 &  &  & \textcolor{gray}{\textbf{3.07}} & \textcolor{gray}{\textbf{213.1}} & \textcolor{gray}{\textbf{1.91}} & \textcolor{gray}{\textbf{324.3}} & \textcolor{gray}{307M} & \textcolor{gray}{64} & \textcolor{gray}{1.0} \\
\hline

\modelname-L-64 &  1.78 & MaskGIT-ViT~\cite{chang2022maskgit} & \textbf{3.64} & 179.8 & 2.74 & 221.1 & \textbf{177M} & \textbf{8} & \textbf{41.0} \\

\multirow{2}{*}{\modelname-B-128} & \multirow{2}{*}{1.37} & \multirow{2}{*}{MaskGIT-ViT~\cite{chang2022maskgit}} & 3.91 & \textbf{182.0} & 2.49 & 260.4 & \multirow{2}{*}{\textbf{177M}} & \textbf{8} & 33.3 \\
& & & 4.17 & 181.0 & \textbf{2.13} & \textbf{261.2} & & 64 & 7.4 \\
\hline

\multicolumn{10}{c}{\textit{ours}} \\
\hline
\texttt{SMAP(KL)-B-128} & 0.69 & \texttt{CARD-B} & 4.29 & 155.2 & 2.99 & 253.1 & 234M & 25 & 11.7 \\
\texttt{SMAP(KL)-B-128} & 0.69 & \texttt{CARD-L} & 3.12 & 211.1 & 2.11 & 298.4 & 568M & 25 & 2.8\\
\bottomrule
\end{tabular}
}
\end{table}

\section{Limitations}
\label{app:limitations}

Our study has several limitations. Most importantly, we evaluate semantic-aware prefix learning only under class-conditional supervision on ImageNet. While this setting is sufficient to isolate the role of semantic conditions in tokenizer pretraining, it remains substantially simpler than text-conditioned or multimodal generation, where semantic inputs are richer and more compositional. In addition, our downstream generation experiments are centered on \texttt{CARD} as a representative generator built on top of \texttt{SMAP}. Although this is adequate to show that semantically grounded tokenization improves generation, broader validation across other generator architectures would be needed to establish full generality. Finally, we restrict our experiments to the image domain. Extending semantic-aware prefix learning to text-conditioned synthesis, multimodal conditioning, and spatiotemporal settings such as video remains an important direction for future work.

\section{Dataset Licenses}
\label{app:dataset_license}
The datasets we used for training and/or testing \texttt{SMAP} are described as follows.

\noindent\textbf{ImageNet-1K:}\quad
We train and evaluate \texttt{SMAP} on ImageNet-1K generation benchmark. This dataset spans 1000 object classes and contains 1,281,167 training images, 50,000 validation images and 100,000 test images. We use the training set for our tokenizer and generator training. The validation set is used to compute reconstruction FID for evaluating tokenizers. The generation results are evaluated with generation FID using pre-computed statistics and scripts from ADM~\cite{dhariwal2021diffusion}~\footnote{\url{https://github.com/openai/guided-diffusion/tree/main/evaluations}}.

License: \url{https://image-net.org/accessagreement}

URL: \url{https://www.image-net.org/}

\end{document}